\documentclass[conference]{IEEEtran}
\IEEEoverridecommandlockouts
\ifCLASSOPTIONcompsoc
  \usepackage[nocompress]{cite}
\else
  \usepackage{cite}
\fi
\ifCLASSINFOpdf
  \usepackage[pdftex]{graphicx}
\else
  \usepackage[dvips]{graphicx}
\fi
\usepackage{amsthm}
\usepackage{multirow} 
\usepackage{amsmath,amssymb,amsfonts}
\usepackage{textcomp}
\usepackage{xcolor}
\usepackage{booktabs}
\usepackage{subcaption} 
\usepackage{diagbox}
\usepackage{float}
\usepackage{makecell} % for \thickhline
\usepackage{array}
\usepackage{svg}
\usepackage{epstopdf}
\usepackage{hyperref}
\newcommand{\orcidlink}[1]{%
  \textsuperscript{\href{https://orcid.org/#1}{\includegraphics[width=1.9ex]{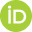}}}%
}
\hypersetup{
    colorlinks=true,
    linkcolor=black,
    citecolor=black,
    urlcolor=black
}
\usepackage[capitalize,noabbrev]{cleveref}
\usepackage{algorithm, algpseudocode}
\usepackage{array}
\usepackage{url}
\theoremstyle{plain}
\theoremstyle{remark}

\newcommand{\parhead}[1]{\vspace{3pt plus 1pt minus 1pt}\par\noindent\textbf{#1}\hspace{.4em plus .2em minus .2em}}
\renewcommand{\paragraph}[1]{\parhead{#1}}

\begin{document}
% ======= delete before sumbit========
% \thispagestyle{plain}
% \pagestyle{plain}

\title{PPFL-RDSN: Privacy-Preserving Federated Learning-based Residual Dense Spatial Networks for Encrypted Lossy Image Reconstruction}

% \author{\IEEEauthorblockN{Peilin He}
% ORCID: 0000-0003-3553-9949 \\
% \IEEEauthorblockA{\textit{Department of Informatics and Networked Systems} \\
% \textit{University of Pittsburgh}\\
% Pittsburgh, USA \\
% peilin.he@pitt.edu}\

% \and
% \IEEEauthorblockN{James Joshi}
% ORCID: 0000-0003-4519-9802\\
% \IEEEauthorblockA{\textit{Department of Informatics and Networked Systems} \\
% \textit{University of Pittsburgh}\\
% Pittsburgh, USA \\
% jjoshi@pitt.edu}

% }
% \markboth{Peilin He \MakeLowercase{\textit{et al.}}: Paper Title Short Version}{}

\author{
Peilin He\,\orcidlink{0000-0003-3553-9949}, James Joshi\,\orcidlink{0000-0003-4519-9802} \\
Department of Informatics and Networked Systems, University of Pittsburgh, Pittsburgh, PA, USA \\
Emails: \{peilin.he, jjoshi\}@pitt.edu
}

% \author{
%     \IEEEauthorblockN{Anonymous Author(s)}
% }

\maketitle
\begin{abstract}
Reconstructing high-quality images from low-resolution inputs using Residual Dense Spatial Networks (RDSNs) is crucial yet challenging. It is even more challenging in centralized training where multiple collaborating parties are involved, as it poses significant privacy risks, including data leakage and inference attacks, as well as high computational and communication costs. We propose a novel Privacy-Preserving Federated Learning-based RDSN (PPFL-RDSN) framework specifically tailored for encrypted lossy image reconstruction. PPFL-RDSN integrates Federated Learning (FL), local differential privacy, and robust model watermarking techniques to ensure that data remains secure on local clients/devices, safeguards privacy-sensitive information, and maintains model authenticity without revealing underlying data. Empirical evaluations show that PPFL-RDSN achieves comparable performance to the state-of-the-art centralized methods while reducing computational burdens, and effectively mitigates security and privacy vulnerabilities, making it a practical solution for secure and privacy-preserving collaborative computer vision applications.
\end{abstract}
\begin{IEEEkeywords}
Privacy-Preserving Federated Learning, Differential Privacy, Watermarking, Lossy Image Reconstruction
\end{IEEEkeywords}

\IEEEpeerreviewmaketitle
\section{Introduction}
\label{1}
Encrypted lossy image reconstruction using neural networks has emerged as a critical area of research to address the need for high-quality image reconstruction while ensuring security, high accuracy, and improved performance \cite{liu2018image,cheng2018deep,akccakaya2022unsupervised}. It involves an image processing pipeline where images are protected before or during compression and later reconstructed from the encrypted, compressed representation by learned models. The aim is to keep content confidential in storage and transit while retaining bitrate savings and the quality gains of neural post-processing, which is valuable for privacy-sensitive, bandwidth-limited application settings such as telemedicine, remote sensing, and cloud photo services. In existing centralized RDSN approaches \cite{peng2022centralized,demelius2023recent,yong2020gradient}, all the training data is expected to be available in the central server before training the model. Thus, centralized training of machine learning (ML) models requires transferring massive amounts of data from participants or data providers to a central server, which creates significant communication overhead, and increases the attack surface for security and privacy compromises \cite{asad2021federated}. This creates substantial scaling challenges in scenarios where there is a large pool of participants/data providers. Furthermore, centralizing data from diverse sources increases the risk of data exposure or leakage, privacy breaches, and malicious or adversarial attacks. To address privacy and concerns and scaling challenges in collaborative ML settings, federated learning (FL) was introduced by McMahan et al. \cite{mcmahan2017communication}. In FL, each participating client independently trains a local ML model using its dataset, thus preventing raw data from being exposed to a central server. After local training, clients send their model updates to an aggregator, which iteratively combines these updates to create a global model. To further protect privacy and maintain the confidentiality of these model updates, various secure and privacy-preserving techniques have been proposed, including Differential Privacy \cite{dwork2006differential, dwork2008differential, dwork2010boosting, dong2022gaussian}, Homomorphic Encryption \cite{yi2014homomorphic, acar2018survey, ogburn2013homomorphic, fan2012somewhat}, and Secure Multi-Party Computation \cite{cramer2015secure, lindell2020secure, goldreich1998secure}. However, implementing these methods introduces additional computational overhead, posing a crucial challenge in balancing privacy guarantees and model accuracy, as well as practically acceptable performance. 

Encryption-then-Lossy-Compression (EtLC) schemes \cite{wang2022novel, zhang2010lossy, hu2014new} have been proposed to securely reconstruct images where two parties are typically involved: a sender and a receiver. EtLC schemes first encrypt an image first and then apply a standard or learned compressor. This ensures that untrusted networks handle only encrypted payloads while staying compatible with off-the-shelf codecs and delivery systems. In practice, EtLC enables secure sharing without sacrificing compression efficiency and pairs naturally with learned reconstructor networks. In these approaches, the receiver must either obtain the Residual Dense Spatial Network (RDSN) model from a third-party, such as a cloud server, or train it independently; however, both approaches have significant drawbacks. Acquiring the model from an untrusted cloud server allows for potential model poisoning attacks, leading to unreliable/corrupted reconstructions at the receiver side. Conversely, local training of the RDSN imposes substantial computational and communication costs. In addition to these practical issues, encrypted image reconstruction tasks face broader challenges of ensuring robust privacy protection and security, especially in scenarios involving multiple distributed data providers, and minimizing the computational burden on individual clients without compromising model performance. Existing RDSN techniques do not address these privacy and security concerns, leaving sensitive information vulnerable during both model training and inference.

To address these critical issues, we propose a novel Privacy-Preserving Federated Learning-based RDSN (PPFL-RDSN) scheme tailored for encrypted lossy image reconstruction in a collaborative, multi-party environment. PPFL-RDSN introduces a federated architecture comprising a central server, responsible for aggregating model updates, and multiple clients collaboratively training an enhanced RDSN model through locally controlled tunable parameters. Building upon and customizing the basic RDSN architecture from Wang et al. \cite{wang2022novel}, our approach enables effective distributed training, achieves high image reconstruction quality while ensuring privacy protection. The proposed PPFL-RDSN framework extends the conventional encryption-then-lossy-compression (EtLC) \cite{chuman2018encryption} paradigm from the traditional two-party setup to a multi-party collaborative setting. Within this extended framework, each client can dynamically assume the roles of sender or receiver, enabling distributed high-resolution image reconstruction without collecting sensitive data from individual providers/participants in a central location. Given that direct integration of differential privacy (DP) noise severely degrades reconstruction quality, our approach leverages Local Differential Privacy (LDP), strategically injecting Gaussian noise into selected feature spaces or parameter updates. This ensures robust privacy without compromising reconstruction quality. Additionally, the proposed mechanism incorporates a watermarking verification mechanism to safeguard model integrity during aggregation.

The key contributions of this paper are as follows:
\begin{itemize}
    \item We propose PPFL-RDSN, a novel federated learning framework that extends encrypted lossy image reconstruction from two-party to multi-party collaboration, providing robust privacy protection.
    \item We introduce an enhanced RDSN architecture tailored for federated learning, which significantly improves image reconstruction quality and reduces training costs.
    \item We develop and employ a Local Differential Privacy approach specifically designed for encrypted image reconstruction, that balances strong privacy protections with minimal accuracy loss.
    \item We integrate a watermarking-based integrity verification mechanism to ensure that clients can reliably verify the authenticity and integrity of aggregated global models.
\end{itemize}
The experimental results show that our approach achieves an accuracy comparable (as measured by PSNR/SSIM metrics) to state-of-the-art methods while improving training efficiency by approximately 6.56\%. More importantly, our framework provides mitigation against inference attacks and model poisoning or alteration attacks, ensuring a balance between utility and privacy in real-world scenarios.
\section{Background and Related Work}
\label{2}
In this section, we overview the key concepts and notation underlying our PPFL\mbox{-}RDSN framework for reconstructing encrypted\mbox{-}then\mbox{-}lossy\mbox{-}compressed images. We first summarize the Residual Dense Spatial Network (RDSN) architecture used for high\mbox{-}fidelity reconstruction, and then revisit federated learning (FL) with a focus on the privacy and integrity mechanisms—most notably differential privacy (DP) and model verification—that are essential when training complex image models across heterogeneous clients.

\subsection{Deep learning-based RDSN}
\label{3}
The Residual Dense Spatial Network (RDSN), proposed by Wang et al.~\cite{wang2022novel}, is a specialized deep\mbox{-}learning architecture designed explicitly for efficiently reconstructing encrypted\mbox{-}then\mbox{-}lossy\mbox{-}compressed images~\cite{goyal2008compressive, al1998lossy, blau2019rethinking}. RDSN extends the Residual Dense Network (RDN) family~\cite{zhang2018residual} by blending two complementary mechanisms: (1) \emph{Residual learning}, which predicts the high\mbox{-}frequency residual between the degraded input and its original counterpart, enabling efficient flow of low\mbox{-}frequency information through skip connections; and (2) \emph{Dense connectivity}, which concatenates feature maps from all preceding layers, encouraging feature reuse and alleviating vanishing\mbox{-}gradient issues. Within each RDN block, the residual path expresses the network’s output as [??]:
\[
\mathbf{y}=\mathcal{F}\left(\mathbf{x},\left\{W_i\right\}\right)+\mathbf{x},
\]
where $\mathbf{x}$ is the input, and $\mathcal{F}$ encapsulates convolutional, activation, and normalization operations parameterized by weights $\left\{W_i\right\}$. Each dense block enhances feature propagation by concatenating the outputs of all preceding layers, as follows:
\[
\mathbf{y}_l = H_l\!\left(\left[\mathbf{x}_0, \mathbf{x}_1, \ldots, \mathbf{x}_{l-1}\right]\right),
\]
where $H_l$ denotes the transformation at layer $l$.

Further, the RDSN model specifically incorporates three crucial enhancements: an improved U\mbox{-}Net architecture~\cite{ronneberger2015}, a Global Skip Connection (GSC), and Uniformly Downsampling Constraints (UDC)~\cite{zhou2020theory}. The refined U\mbox{-}Net employs a sub\mbox{-}pixel convolution operation, defined by $\mathbf{Y}=\operatorname{PS}(\mathbf{W} * \mathbf{I})$, where $\mathbf{W}$ represents the convolutional kernel, $*$ denotes convolution, and $\operatorname{PS}(\cdot)$ performs the pixel shuffle operation that rearranges elements to enhance spatial resolution. The Global Skip Connection significantly improves efficiency by directly propagating essential features from input to deeper layers, preserving crucial structural information:
\[
\mathbf{I}_{\text{out}} = \mathbf{F}_{\mathrm{RDSN}}\!\left(\mathbf{I}_{\mathrm{in}};\theta\right)+\mathbf{I}_{\mathrm{in}},
\]
where $\mathbf{I}_{\text{in}}$ and $\mathbf{I}_{\text{out}}$ denote the input and output images, respectively, and $\theta$ represents network parameters. Additionally, the integration of Uniformly Downsampling Constraints ensures the retention of essential spatial features during the reconstruction of downsampled images. Given an input image $\mathbf{I}_{\text{in}}$ and a scaling factor $s$, the uniformly downsampled output is:
\[
\mathbf{I}_{\mathrm{UD}}=\text{Downsample}\!\left(\mathbf{I}_{\mathrm{in}}, s\right).
\]

While encryption\mbox{-}then\mbox{-}lossy\mbox{-}compression pipelines have been explored to prevent direct content inspection~\cite{chuman2018encryption, kurihara2015encryption}, these prior solutions typically assume a two\mbox{-}party setting with a trusted receiver. Scaling to multi\mbox{-}participant collaboration introduces coordination and trust challenges around both the encrypted data and the global model’s authenticity. The closest existing work related to our proposed approach is the RDSN approach proposed in~\cite{wang2022novel}, which targets reconstruction quality but does not integrate federated contrastive training~\cite{li2020federated}, enhanced privacy protections, or cryptographic verification. Our framework operates in this distributed, multi\mbox{-}party setting, where an acceptable, higher reconstruction quality must be preserved while providing privacy and integrity guarantees.

\subsection{Federated Learning}
\label{5}
Federated Learning (FL) enables multiple clients to collaboratively train a global model by locally updating on their private data and sharing updates with a central aggregator, thereby reducing raw\mbox{-}data exposure~\cite{mcmahan2017communication}. Concretely, each client or device $k$ performs local optimization
\[
w_k^{(t+1)} = w_k^{(t)} - \eta \nabla F_k\!\big(w_k^{(t)}, \mathcal{B}_k\big),
\]
where $w_k^{(t)}$ are the model parameters of device $k$ at iteration $t$, $\eta$ is the learning rate, and $\nabla F_k(\cdot,\mathcal{B}_k)$ is the gradient computed on local mini\mbox{-}batch $\mathcal{B}_k$. The aggregator then forms the new global model via weighted averaging (FedAvg \cite{li2019convergence}):
\[
w^{(t+1)} = \sum_{k=1}^{K} \frac{n_k}{n}\, w_k^{(t+1)},
\]
where $K$ is the number of participating devices, $n_k$ is the number of data points on device $k$, and $n=\sum_{k=1}^{K} n_k$. This iterative averaging reduces the number of communication rounds relative to synchronized SGD~\cite{chen2016revisiting, lian2018asynchronous}, and has been deployed across diverse domains such as image processing, large language models, and explainable AI—where data locality and heterogeneity are the norm~\cite{briggs2020federated, buyukates2021timely, mammen2021federated, bagdasaryan2020backdoor, zhang2021survey, konecny2016federated, zhu2021federated, lyu2020threats, chen2021communication, khan2021federated}.

However, two considerations become critical for high\mbox{-}fidelity image reconstruction in FL. \emph{First}, most of the existing FL work for image analysis has focused on comparatively simpler objectives such as classification or segmentation~\cite{feng2022specificity, elmas2022federated, guo2021multi}. Reconstruction demands preserving subtle textures and high\mbox{-}frequency details under distribution shift and client heterogeneity, which stresses both the optimization and the communication related to computational costs. \emph{Second}, although FL reduces direct data sharing, it does not by itself prevent privacy leakage through shared updates, nor does it ensure protection against malicious model tampering during the iterative aggregation process.

To address privacy leakage, differential privacy (DP)~\cite{dwork2006differential, dwork2008differential} offers formal guarantees that the released outputs (e.g., clipped\mbox{+}noise-added gradients or differentially private final model) do not reveal sensitive information about any individual record. In practice, however, naively injecting noise can adversely impact accuracy in high\mbox{-}dimensional images, a phenomenon particularly problematic for reconstruction quality. Recent work explores refined mechanisms that adapt noise to internal model states to reduce utility loss~\cite{wang2024dpadapter, li2024fine, phan2017adaptive}; yet these often introduce deployment overheads or rely on trust assumptions about the execution environment. Moreover, practical challenges — including floating\mbox{-}point implementation pitfalls~\cite{holohan2021secure, mironov2012significance}, precision\mbox{-}based attacks~\cite{haney2022precision}, and erroneous estimation of sensitivity~\cite{casacuberta2022widespread}—can undermine DP guarantees if the DP solution is not properly designed. These challenges are amplified in image reconstruction tasks, where excessive noise can manifest as blur and loss of detail. Our proposed design is therefore directed toward the development of privacy mechanisms and training protocols that safeguard fine-grained FL-related structural information while ensuring stringent privacy guarantees.

In addition to privacy, we need appropriate integrity verification approaches to ensure that the global model has neither been replaced nor maliciously altered across training rounds. Watermarking and fingerprinting techniques~\cite{papernot2017practical} provide one way to authenticate the trained model and detect unauthorized modifications, allowing clients to verify the identity and authenticity of the downloaded global models before use. In higher\mbox{-}assurance settings, cryptographic protocols such as zero\mbox{-}knowledge proofs~\cite{fiat1986prove} can further attest to correct aggregation without exposing private updates. Together with FL\mbox{-}appropriate learning objectives (e.g., federated contrastive training~\cite{li2020federated}), these mechanisms can help establish a training loop in which neither raw data nor model parameters can be exploited for information leakage or manipulation, while still meeting the fidelity demands of complex reconstruction.
\begin{figure*}[t]
    \centering
    \includegraphics[width=\linewidth, height=9.5cm]{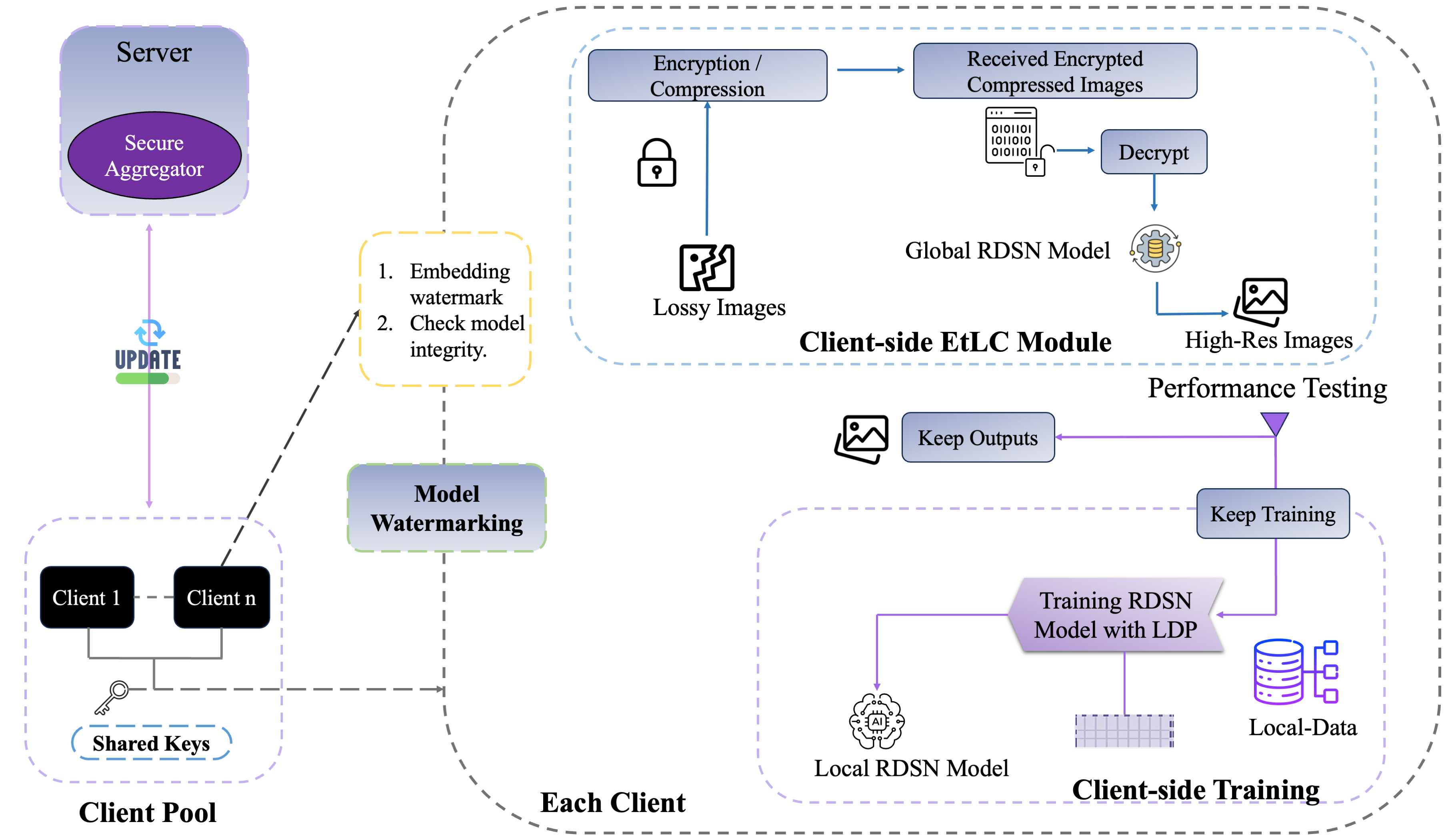}
    \caption{An Overview of the PPFL-RDSN Framework}
    \label{fig:PPFL-RDSN}
\end{figure*}
\section{The Proposed PPFL-RDSN Framework}
\label{Proposed PPFL-RDSN Framework}
In this section, we overview our proposed PPFL-RDSN framework, illustrated in \textbf{\cref{fig:PPFL-RDSN}}: 
\begin{itemize}
    \item \textbf{Local Training}: Within the PPFL-RDSN framework, each client carries out its local model training, maintaining exclusive access to its dataset without sharing it with other clients. During local training, Gaussian noise is strategically applied to the parameter updates to balance data utility and privacy. Additionally, each client embeds a watermark (one time) into its local RDSN model during training for subsequent integrity verification.
    \item \textbf{Model Aggregation and Distribution}: 
    After training, clients securely transmit their local models to a central server that acts as a secure aggregator. The server performs model aggregation using Federated Averaging (FedAvg) \cite{li2019convergence}, combining local models into a global model based on weighted contributions.
    \item \textbf{Local EtLC}: 
    Once aggregated, the global model is distributed back to each client. Upon receipt, clients execute a model integrity check using their embedded watermark. After successful integrity verification, we input lossy or low-resolution images into the global model and carry out performance testing. Depending on the testing outcomes, the client either retains the global model or further initiates the next iteration of local training. Additionally, the Local EtLC subsystem allows clients to securely exchange encrypted and compressed low-resolution images. We assume pre-established shared secret keys for encrypted communication. Upon request, these images are decrypted, decompressed, and reconstructed into higher-resolution outputs using the verified global RDSN model. This process ensures collaborative enhancement while maintaining data privacy and integrity throughout the federated learning process.
\end{itemize}

\subsection{Threat Modeling and Mitigation}
\label{threat_model}
We analyze three classes of adversaries: 1) a honest-but-curious aggregator, 2) honest‑but‑curious clients (who may collude for inference while still following the protocol), and 3) a passive network eavesdropper. We focus the threat model on attacks that are countered by our current design.

\textbf{Privacy Threats:} 
    Membership, inversion, and property inference attacks exploit shared gradients to uncover private training data \cite{shokri2017membershipinferenceattacksmachine, fredrikson2015model, ganju2018property}. These threats arise from both untrusted or semi-trusted aggregators as well as honest-but-curious clients that may pool their local observations and round-wise global models. Since our mechanism satisfies local $(\varepsilon,\delta)$-differential privacy, its guarantee is robust to any arbitrary side information—including potential collusion among clients or the aggregator because of the post-processing and composition properties of differential privacy \cite{dwork2014algorithmic, dwork2010boosting, kairouz2021advances}.

\textbf{Security Threats:} 
    Model substitution or parameter tampering attacks \cite{papernot2016transferability, papernot2017practical} utilize the untrusted server to replace the global model with a modified tampered version;  rollback or replays of old models form a related threat. Passive eavesdropping in transport can observe traffic, but cannot bypass end-to-end integrity checks; any active modification reduces to the substitution/tampering case. In our design, we do not claim our mechanism has the ability to defend against the following poisoning and backdoor attacks \cite{bagdasaryan2019backdoorfederatedlearning}.

\textbf{Mitigation Strategies}: To address these threats, we integrate the following protection mechanisms:
    \begin{itemize}
        \item \textit{Local Differential Privacy}: Our carefully tailored additive DP noise is applied to high-frequency DCT components of client activations/updates with clipping-based sensitivity control, preventing the inference attacks above by the aggregator or colluding honest-but-curious clients. Per-round $(\varepsilon,\delta)$-LDP composes across rounds via the moments accountant. The detailed design of LDP is described in \cref{local-dp}.
        \item \textit{Model Watermarking Mechanism}: As described in \cref{watermark}, a sparse PRF-keyed watermark is embedded during client training and verified by clients \emph{before} local use of the aggregated model. This prevents model substitution or unauthorized alterations (including replay/rollback) by the aggregator or an active on-path adversary, ensuring that global updates are accepted only if provenance and integrity are verified.
    \end{itemize}

\subsection{Local Differential Privacy (LDP)}
\label{local-dp}
Building on the standard LDP framework \cite{arachchige2019local, truex2020ldp}, we inject noise inside the client model at feature domains where perturbations least harm utility. Unlike prior frequency‑domain privacy schemes designed for face recognition \cite{ji2022privacy, arachchige2020privacy}, our LDP is fully differentiable and thus fits end‑to‑end reconstruction networks.

\paragraph{Step\,1: Data-independent orthogonal transform.}
For a fixed layer~$k$, denote its activation
\[
  H = g_k(\dots, X; W_k) \in \mathbb{R}^{C \times h \times w},
\]
where $g_k$ with parameters $W_k$ produces a tensor with $C$ channels and spatial size $h \times w$. We apply a data-independent orthogonal transform per channel:
\[
  C_c \;=\; T H_c T^{\!\top}\quad\text{for }c=1,\dots,C,
\]
where $H_c \in \mathbb{R}^{h \times w}$ is the $c$-th slice and $T\in\mathbb{R}^{h\times h}$ is the (orthonormal) 2-D Discrete Cosine Transform (DCT) matrix along each spatial axis.\footnote{Equivalently,
$\operatorname{vec}(C_c)=(T\otimes T)\operatorname{vec}(H_c)$ with $T\otimes T$ orthonormal.}
Because $T$ is orthonormal, the transform is an isometry:
\[
  \|C_c\|_F=\|H_c\|_F \quad\text{and}\quad \|C_c\|_2=\|H_c\|_2,
\]
and thus, for any $\ell_2$-based clipping or sensitivity measure that is invariant under orthogonal changes of basis, the sensitivity is unchanged in the transform domain. This norm preservation lets us calibrate DP noise in the DCT domain exactly as in the original domain.

\paragraph{Step\,2: Deterministic low/high‑frequency partition.}
Let $(u,v)$ index the 2‑D spectrum of~$C$.  
We pre‑define a frequency threshold $\tau$ (e.g., $\tau=8$ in DCT order):
\[
  \mathcal{S}_L=\{(u,v)\mid u+v<\tau\},\quad
  \mathcal{S}_H=\{(u,v)\mid(u,v)\notin\mathcal{S}_L\}.
\]
Because the frequency threshold~$\tau$ is fixed a priori and independent of private data, the partition is data-independent. By the post-processing invariance of differential privacy, deterministic operations that do not depend on private inputs incur no additional privacy loss \cite{dwork2014algorithmic}.

We retain the coefficients in $\mathcal{S}_L$ (capturing global structure) and add noise to those in $\mathcal{S}_H$ (capturing fine detail).  With
$C_L$ and $C_H$ as the masked tensors, we get,
$H_L=T^{\!\top}C_LT$, $H_H=T^{\!\top}C_HT$.

\paragraph{Step\,3: Clipping and sensitivity.}
We clip $H$ by an $L_2$ norm~$C_{\max}$:
\[
  \widehat{H}=H/\max\!\bigl(1,\|H\|_2/C_{\max}\bigr).
\]
For neighbouring inputs $X,X'$ that differ in one sample,
the clipped activation change satisfies
$\|\widehat{H}(X)-\widehat{H}(X')\|_2\le 2C_{\max}$.
Because $T$ is orthogonal,
the same bound holds for the stacked vector
$f(X)=\operatorname{vec}(T\widehat{H}T^{\!\top})_{\mathcal{S}_H}$,
so the global $L_2$ sensitivity is
\begin{equation}
  \Delta_2 f = 2C_{\max}.
  \label{eq:sensitivity}
\end{equation}

\paragraph{Step\,4: Gaussian mechanism on~$C_H$.}
We sample independent Gaussian noise and reshape it back to the spectrum:
\[
  \widetilde{f}=f(X)+\mathcal{N}\!\bigl(0,\sigma^2 I\bigr),\qquad
  \sigma=\frac{\Delta_2f\sqrt{2\ln(1.25/\delta)}}{\epsilon}.
\]
By the Gaussian mechanism \cite{dwork2014algorithmic} 
the released $\widetilde{f}$ is $(\epsilon,\delta)$‑DP.
The noisy activation is reconstructed as follows:
\begin{equation}
  \widetilde{H} \;=\; H_L \;+\;
  T^{\!\top}\bigl(\operatorname{reshape}(\widetilde{f})\bigr)T.
  \label{eq:noisyH}
\end{equation}

\paragraph{Algorithm 1: LDP-ClientUpdate.}
The full client routine is summarized in \cref{alg:cldp}.
Multiple rounds are composed via the moments accountant
in our experiments, $30$ local steps with $\sigma{=}0.8$
yield a total $\epsilon\!\approx\!3$ at $\delta=10^{-5}$.
\begin{algorithm}[t]
\caption{\textsc{LDP‑ClientUpdate}}
\label{alg:cldp}
\begin{algorithmic}[1]
\Require local batch $B$, model $W$, layer $k$, clip $C_{\max}$,
         DP parameters $(\epsilon,\delta)$
\State $H\gets g_k(\dots,X;W_k)$ \Comment{Forward to layer $k$}
\State $\widehat{H}\gets \textsc{ClipL2}(H,C_{\max})$
\State $C\gets T\widehat{H}T^{\!\top}$  \Comment{Fixed orthogonal transform}
\State $(C_L,C_H)\gets\textsc{Partition}(C,\tau)$
\State $f\gets\operatorname{vec}(C_H)$
\State $\sigma\gets (2C_{\max}\sqrt{2\ln(1.25/\delta)})/\epsilon$
\State $\widetilde{f}\gets f+\mathcal{N}(0,\sigma^2 I)$
\State Re‑shape $\widetilde{f}$ into $C_H^{\text{noise}}$
\State $\widetilde{H}\gets H_L+T^{\!\top}C_H^{\text{noise}}T$
\State Continue forward/back‑prop with $\widetilde{H}$
\State Send $\Delta\widetilde{W}_i$ (or $\widetilde{H}$) to server
\end{algorithmic}
\end{algorithm}

In summary, our client‑side LDP module safeguards the input data while retaining reconstruction quality: we first partition DCT‑domain activations with a publicly fixed threshold, incurring no privacy cost or an explicitly budgeted one if adaptively learned; we then clip the activations and exploit the norm‑preserving orthogonality of the DCT to derive a tight global $L_2$ sensitivity of $2C_{\max}$; finally, Gaussian noise calibrated to this bound delivers formal $(\epsilon,\delta)$‑DP guarantees, with privacy loss composed across rounds via the moments accountant. Because noise is confined to high-frequency coefficients that contribute less to overall perceptual fidelity, the dominant low-frequency content remains largely intact, enabling strong privacy protection with minimal degradation in reconstruction quality within the PPFL-RDSN framework.

\subsection{Model Watermarking}
\label{watermark}

To ensure provenance and integrity of local updates in the PPFL--RDSN framework, we embed a client‑specific digital watermark that (i) does not degrade task performance, (ii) survives federated aggregation, and (iii) remains detectable even in very high‑dimensional models. Our design builds upon two prior approaches: parameter-regularizerr watermarking \cite{Uchida_2017} and double-layer optimization for robustness \cite{yang2021robust}. Empirically, such schemes incur $<0.05\%$ accuracy loss while withstanding pruning of $65\%$ weights or 10 epochs of fine‑tuning, we show detailed results in \cref{tab:wm_ablation}. 

\paragraph{Watermark key and carrier subset.}
Parameter regularization–based watermarking has been extensively studied for centralized neural networks, where a small subset of parameters is regularized toward secret values during training \cite{uchida2017embedding, boenisch2021systematic}. Such methods, however, rely on a single-owner model and independently chosen watermark locations, making them incompatible with federated learning where parameter updates are aggregated across multiple clients. 
Recent works on federated watermarking \cite{li2022fedipr} still treat each client’s watermark as isolated, which limits detectability and robustness once the updates are mixed.

To address these limitations, we propose a coordinated sparse model watermark mechanism: 
all clients share a secret Pseudorandom Function (PRF) key $mk$ to deterministically select an identical, extremely sparse carrier subset $J \subset [d]$, and embed tiny regularization biases only on those coordinates. This design is new in that (i) it preserves client-side privacy while (ii) amplifying watermark detectability at the global-aggregation level, and (iii) maintains indistinguishability from random noise at the single-client level. 
Below, we formalize how $J$ is sampled and why such coordination remains secure.

Let $\text{PRF}(mk,x)$ be a secure pseudorandom function.  
We first deterministically sample a small, fixed index set:
\[
    J=\{j\mid\text{PRF}(mk,j)<p\}, \quad |J| = p d\ll d ,
\]
where $d$ is the model dimension and $p\!\in\!(10^{-4} - 10^{-2})$.  
We reuse the same secret key $mk$, which makes every client perturb an identical, extremely sparse coordinate subset $J$ chosen by a PRF, so their updates reinforce each other and the watermark emerges clearly only in the aggregated model. This coordination remains secure because the subset $J$ is generated through a secret-keyed pseudorandom function (PRF). 
Without the key $mk$, an adversary cannot distinguish whether a given index $j$ was intentionally selected or randomly chosen—the distribution of selected indices is computationally indistinguishable from uniform random sampling. Choosing a small sparsity level $p$ limits the number of perturbed coordinates, allowing each client’s per-coordinate watermark amplitude $\alpha$ to be set far below the standard deviation of stochastic gradients $\sigma$. Consequently, the client-side signal-to-noise ratio $\mathrm{SNR}_{\text{client}}=\alpha^2/\sigma^2$ remains $\ll 1$, meaning that an individual watermark update is statistically indistinguishable from ordinary gradient noise.
Because each client’s watermark perturbation is extremely weak ($\mathrm{SNR}_{\text{client}}\!\ll\!1$) and affects only a $p$-fraction of coordinates, identifying the true carrier set $J$ from noisy updates amounts to a high-dimensional sparse-support recovery problem. Hence, $J$ remains effectively hidden even against large collaborative attacks. Thus, sharing the same coordinates is a deliberate design that maximizes watermark detectability while preserving secrecy and robustness.

\paragraph{Training‑time embedding via regularisation.}
Instead of post‑hoc noise injection, we jointly optimise the task loss $\mathcal{L}_{\text{task}}$ and a watermark loss
$
    \mathcal{L}_{\text{wm}}
        = \frac{1}{|J|} \sum_{j\in J} (\theta_j w_j - \beta)^2, 
$
$
    \text{with }    
    w_j \!=\! \text{PRF}(mk,(w,j)), \;
    \beta>0,
$
which drives the correlation between $\theta_j$ and $w_j$ towards the target $\beta$.  
The total objective on each client is
\(
    \mathcal{L}=\mathcal{L}_{\text{task}}+\lambda \mathcal{L}_{\text{wm}},
\)
where $\lambda = 10^{-4}$ is tuned so that validation accuracy drop stays below 0.1\%.  
Because $\mathcal{L}_{\text{wm}}$ is quadratic and only acts on $|J|/d$ parameters, its contribution to total gradient norm is bounded by $\lambda\beta^2 p$, guaranteeing negligible drift of the optimisation trajectory.

\paragraph{Statistical verification.}
Let \(J=\operatorname{PRF}_{mk}(\text{seed})\subseteq[d]\) be the watermark carrier of size \(|J|\) and
\(w\in\{\pm\beta\}^{|J|}\) its signed pattern.  
After training, the auditor (here refers to each client) computes the test statistic
\[
    S=\frac1{|J|}\sum_{j\in J}\theta_j\,w_j,
    \qquad\text{where }\theta\in\mathbb{R}^{d}\text{ is the released model.}
\]

\emph{Null hypothesis \(\boldsymbol{H_0}\) (no watermark).}  
If the model is unmarked, \(\{\theta_j\}_{j\in J}\) and \(w\) are independent.
Assume the products \(X_j:=\theta_jw_j\) are independent, zero-mean, and bounded,
\(|X_j|\le\sigma\).  Hoeffding’s inequality gives
\[
    \Pr\!\bigl[S>\delta\bigr]
    =\Pr\!\Bigl[\frac1{|J|}\sum_{j=1}^{|J|}X_j>\delta\Bigr]
    \le\exp\!\Bigl(-\tfrac{2|J|\delta^{2}}{\sigma^{2}}\Bigr).
\]

\emph{Alternative hypothesis \(\boldsymbol{H_1}\) (watermark present).}  
With watermark strength \(\beta>0\), each client nudges coordinate \(j\in J\) by at least \(w_j\), so
\[
    \mathbb{E}[X_j]\ge \beta-\lambda,
\]
where \(\lambda\) upper-bounds any adversarial bias on \(\theta_j\).  Concentration again yields
\[
    \Pr\!\bigl[S<\tfrac12\beta\bigr]
    \le\exp\!\Bigl(-\tfrac{2|J|(\beta/2-\lambda)^{2}}{\sigma^{2}}\Bigr).
\]

\emph{Parameter choice.}  
Choosing \(\delta=\beta/2\), \(\lambda\ll\beta\), and \(|J|\ge500\) gives
\[
    \Pr_{H_0}[S>\beta/2] < 10^{-6},
    \qquad
    \Pr_{H_1}[S\le\beta/2] < 10^{-3},
\]

\paragraph{Algorithm 2: PPFL Watermark.}
\cref{alg:wm} outlines both the client-side embedding and the server-side verification procedures.  
Each client deterministically selects an identical sparse index set $J=\{j\mid \text{PRF}(mk,j)<p\}$ using a shared secret key $mk$, and associates each coordinate $j\!\in\!J$ with a pseudorandom sign $w_j=\text{PRF}(mk,("w",j))$.  
During local training, clients add a small regularization term driving $\theta_j w_j$ toward a target value $\beta$, ensuring that all clients perturb the same coordinates in the same (signed) direction.  
Because these updates are aligned across clients, the watermark components reinforce each other when aggregated under FedAvg or its variants, rather than cancel out.  
In verification, the auditor recomputes $J$ and $w_j$ from $mk$, evaluates the normalized correlation score $\frac{1}{|J|}\sum_{j\in J}\theta_j w_j$, and declares a valid watermark if the score exceeds a detection threshold $\delta$.

\begin{algorithm}[t]
\caption{PPFL Watermark: \textsc{Embed \& Verify}}
\label{alg:wm}
\begin{algorithmic}[1]
\Require Model weights $\theta\!\in\!\mathbb{R}^d$, dataset $\mathcal{D}$ (client), secret key $mk$, prob.\ $p$, reg.\ weight $\lambda$, target $\beta$, threshold $\delta$, learning‑rate $\eta$
% ---------- Client side ----------
\Procedure{ClientTrain}{$\theta, \mathcal{D}, mk, p, \lambda, \beta, \eta$}
    \State $J\gets\{j\mid\text{PRF}(mk,j)<p\}$;\quad
           $w_j\gets\text{PRF}(mk,("w",j))$ for $j\!\in\!J$
    \ForAll{mini‑batch $(x,y)\sim\mathcal{D}$}
        \State $\mathcal{L}\gets\mathcal{L}_{\text{task}}(x,y;\theta)
                 +\lambda\frac1{|J|}\sum_{j\in J}(\theta_j w_j-\beta)^2$
        \State $\theta\gets\theta-\eta\nabla_\theta\mathcal{L}$
    \EndFor
    \State \Return $\theta$ \Comment{upload to server}
\EndProcedure
% ---------- Server / auditor ----------
\Procedure{VerifyWatermark}{$\theta, mk, p, \delta$}
    \State $J\gets\{j\mid\text{PRF}(mk,j)<p\}$;\quad
           $w_j\gets\text{PRF}(mk,("w",j))$ for $j\!\in\!J$
    \State $\text{score}\gets\frac1{|J|}\sum_{j\in J}\theta_j w_j$
    \State \Return $(\text{score}>\delta)$
\EndProcedure
\end{algorithmic}
\end{algorithm}

\section{Experimental Evaluation and Analysis}
\label{sec:exp}
In this section, we present an end‑to‑end implementation of our proposed PPFL‑RDSN, and analyze how different configuration choices affect its efficiency, model quality, and privacy guarantees. Our evaluation targets two complementary questions:
\begin{enumerate}
    \item \textbf{Accuracy and Robustness}. Does the framework reach state‑of‑the‑art super‑resolution accuracy on established image benchmarks?  
    \item \textbf{Scalability and Privacy}. How does performance change as we (i) vary the number of clients and training rounds, (ii) inject differential privacy (DP), and (iii) enable watermark‑based model verification?  
\end{enumerate}
Each experiment is mapped to one of the above goals and adopts standard super‑resolution metrics (PSNR and SSIM) and the same data splits as prior works \cite{wang2022novel, mirț2022downsampling, zhang2018residual}.

\subsection{Experimental Setup}
\label{subsec:setup}
We deploy our system in the simulation environment. This system is designed on the local machine, and each client is simulated as a multi-threaded processor. 

\textbf{(I) Simulation platform}.  All code is written in \texttt{Python~3.10} with \texttt{PyTorch~1.12.8} and the CUDA~11.2 toolkit. A single Fedora/Red Hat Linux host emulates the central server and all clients; each client runs in a separate thread and is bound to its own NVIDIA~L40 GPU. Opacus~\cite{opacus} is modified to provide record‑level DP.

\textbf{(II) Datasets}.  Following Wang et al.~\cite{wang2022novel}, we use DIV2K, Set5, Set14, BSDS100, and Urban100. DIV2K’s 800 high‑resolution (HR) images and their down‑sampled low‑resolution (LR) counterparts (scaling factors $\varsigma\!\in\!\{2,3,4\}$) are randomly partitioned into disjoint local subsets $\mathcal{D}_{i}$ that never leave their respective clients.  The remaining HR/LR pairs constitute the global validation and test sets.

\textbf{(III) Hyper‑parameters}.  If a client lacks a pretrained model, it first initializes $\theta_{0}$ and performs local training.  Local RDSN uses 1000 epochs (as in~\cite{wang2022novel}) with Adam optimizer \cite{diederik2014adam}, batch size~16, $64^{2}$ LR crops, and an exponentially decaying learning rate starting at $1\times10^{-4}$.  In FL, 10~clients each runs $x$ local epochs per round over 10~global rounds, for a total of $10x$ epochs.  The overall training time in an ideal environment is:
\[
  \!\big[T_{\text{total}}(\varepsilon)\big]
  \;\approx\;
  [R_\varepsilon]\,
  \Big(
    [T_{c,\max}] + [T_{\text{agg}}] + [T_{\text{veri}}]
  \Big)
  \;+\; T_{\text{init}} \;+\; T_{\text{final}}
\]

where $
T_{c,\max}\;:=\;\max_{i\in S}\Big(T^{\downarrow}_{i}+T^{\text{updates}}_{i}+T^{\uparrow}_{i}\Big)
$, which is the slowest client’s time, $T_{\text{veri}}$ (watermark verification) is negligible for $<\!10\,000$ clients, and all other symbols follow Juan et al.~\cite{Juan2024EnhancingCE}.

\textbf{(IV) Baseline timing}.  Re‑running Wang \emph{et al.} with two NVIDIA~L40 GPUs took 82,800s for 1000 centralized epochs. Our PPFL‑RDSN converges in only 47 local epochs per round (470 total), cutting wall‑clock time to 77,368s while additionally guaranteeing privacy.

\textbf{(V) Evaluation metrics}.  Images are assessed with peak signal‑to‑noise ratio (PSNR) and structural similarity (SSIM); higher values indicate better reconstructions.

\subsection{Key Management for the Local EtLC Module}
For the key sharing in the Local EtLC module, shown in \cref{fig:PPFL-RDSN}, we adopt a standards‑based design that keeps the aggregator untrusted and adds minimal overhead. Each EtLC object is encrypted under a fresh content‑encryption key (CEK) with an Authenticated Encryption with Associated Data (ChaCha20‑Poly1305) \cite{nir2018chacha20}, and the CEK is delivered to authorized recipients using Hybrid Public Key Encryption (HPKE) \cite{RFC9180, RFC9420, X3DH}. First contact between previously unknown peers can follow Extended Triple Diffie–Hellman to authenticate identity keys and obtain forward secrecy; for 1 to n or dynamic recipients, Messaging Layer Security (MLS) \cite{BBS98} maintains a group secret with forward secrecy and post‑compromise security, from which per‑image CEKs are derived. In scenarios where encrypted EtLC content must later be made accessible to additional clients, proxy re-encryption offers a practical approach: it enables ciphertexts of content-encryption keys to be re-targeted to new recipients by an untrusted proxy, while preserving confidentiality of both the keys and the underlying data. These mechanisms are orthogonal to LDP and watermarking, preserving our training and verification pipeline while providing robust key management.

\subsection{Performance Evaluation}
\label{subsec:eval}
We adopt the standard super‑resolution benchmarks: Set5, Set14, BSDS100, and Urban100, and the three common scaling factors \(\varsigma\in\{2,3,4\}\).
Following 470 global epochs on randomly partitioned DIV2K shards, the latest \emph{vanilla} FL‑RDSN model (no differential privacy~(DP) and no watermarking) is broadcast to a dedicated evaluation client that first applies the encryption‑then‑lossy‑compression pipeline before inference. We use
PSNR and SSIM as primary quality metrics. The comparison results are shown in Appendix Sec. \ref{comparison}.

% ============================table1====================
\begin{table}[t]
\centering
\caption{\textit{Performance evaluation using PSNR and SSIM standard metrics for 4 different datasets, 3 different scaling factors for our FL-RDSN model, and comparison with the State-of-the-Art.}}
\Large 
\renewcommand{\arraystretch}{1.3}
\resizebox{\columnwidth}{!}{
\begin{tabular}{ccccccccc}
\toprule
Datasets    & \multicolumn{2}{c}{Set5}        & \multicolumn{2}{c}{Set14}       & \multicolumn{2}{c}{BSDS100} & \multicolumn{2}{c}{Urban100}    \\ \hline
\multicolumn{9}{c}{\textbf{\(\varsigma\) = 2}}                                                                                    \\ \hline
Metrics     & PSNR  & SSIM           & PSNR           & SSIM  & PSNR           & SSIM           & PSNR           & SSIM           \\ \hline
Wang et al. & 39.45 & \textbf{0.972} & 35.06          & 0.943 & 33.53          & 0.931          & 33.84          & 0.952          \\
FL-RDSN     & 39.45          & 0.964          & \textbf{35.13} & 0.934          & \textbf{33.56}    & 0.931   & \textbf{33.95} & \textbf{0.953} \\ \hline
\multicolumn{9}{c}{\textbf{\(\varsigma\) = 3}}                                                                                    \\ \hline
Wang et al. & \textbf{35.17} & 0.939          & 31.01          & \textbf{0.870} & 29.75             & 0.842   & \textbf{29.04} & 0.882          \\
FL-RDSN     & 35.15 & 0.939          & \textbf{31.03} & 0.869 & 29.75          & \textbf{0.843} & 29.03          & \textbf{0.884} \\ \hline
\multicolumn{9}{c}{\textbf{\(\varsigma\) = 4}}                                                                                    \\ \hline
Wang et al. & \textbf{32.72} & \textbf{0.904} & \textbf{29.06} & \textbf{0.804} & 27.95             & 0.764   & 26.63          & 0.814          \\
FL-RDSN     & 32.69 & 0.894          & 29.05          & 0.799 & \textbf{27.96} & \textbf{0.765} & \textbf{26.72} & 0.814          \\ \hline
\end{tabular}
}
\label{FL-benchmark}
\end{table}
% =========================table1end===================
% ==========================table 2============= 
\begin{table}[t]
\centering
\caption{\textit{Evaluation of Vanilla FL-RDSN vs PPFL-RDSN using 4 different datasets and 3 different scaling factors.}}
\Large
\renewcommand{\arraystretch}{1.3}
\resizebox{\columnwidth}{!}{%
\begin{tabular}{ccccccccc}
\hline
\multicolumn{9}{c}{$\varepsilon$ = 2.75}                     \\ \hline
Datasets & \multicolumn{2}{c}{Set5} & \multicolumn{2}{c}{Set14} & \multicolumn{2}{c}{BSDS100} & \multicolumn{2}{c}{Urban 100} \\ \hline
\multicolumn{9}{c}{\textbf{\(\varsigma\) = 2}}                            \\ \hline
Metrics   & PSNR  & SSIM  & PSNR  & SSIM  & PSNR  & SSIM  & PSNR  & SSIM  \\ \hline
FL-RDSN   & 39.45 & 0.964 & 35.13 & 0.934 & 33.56 & 0.931 & 33.95 & 0.953 \\
PPFL-RDSN & 39.17 & 0.963 & 34.96 & 0.932 & 33.45 & 0.930 & 33.56 & 0.948 \\ \hline
\multicolumn{9}{c}{\textbf{\(\varsigma\) = 3}}                            \\ \hline
FL-RDSN   & 35.15 & 0.939 & 31.03 & 0.869 & 29.75 & 0.843 & 29.03 & 0.884 \\
PPFL-RDSN & 34.99 & 0.925 & 30.97 & 0.855 & 29.70 & 0.838 & 28.84 & 0.874 \\ \hline
\multicolumn{9}{c}{\textbf{\(\varsigma\) = 4}}                            \\ \hline
FL-RDSN   & 32.69 & 0.894 & 29.05 & 0.799 & 27.96 & 0.765 & 26.72 & 0.814 \\
PPFL-RDSN & 32.52 & 0.884 & 28.97 & 0.785 & 27.91 & 0.758 & 26.52 & 0.802 \\ \hline
\end{tabular}
}
\label{dp-comparison}
\end{table}
% ==================end================

\paragraph{Baseline Performance.}
\label{subsec:baseline}
\cref{FL-benchmark} compares our vanilla FL‑RDSN with the centralized model of Wang~\emph{et~al.}  They are statistically indistinguishable across all benchmarks. For larger or diverse datasets, FL‑RDSN is occasionally \emph{better}.  \cref{fig:psnr-comparison,fig:ssim-comparison} show per‑image results showing that the aggregate scores reflect consistent image‑level behavior.  

\begin{figure}[t!]
    \centering
    % 第一行：PSNR
    \begin{subfigure}[b]{0.49\linewidth}
        \includegraphics[width=\linewidth, height=4cm]{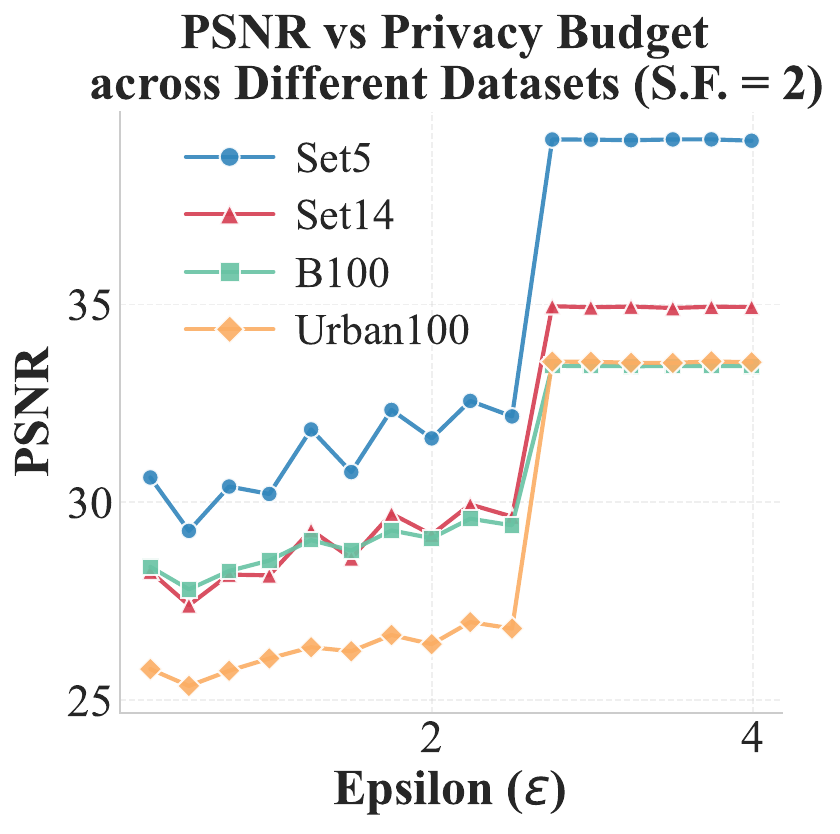}
        \caption{$\epsilon$ to PSNR, $\varsigma = 2$}
        \label{fig:sub1}
    \end{subfigure}
    \begin{subfigure}[b]{0.49\linewidth}
        \includegraphics[width=\linewidth]{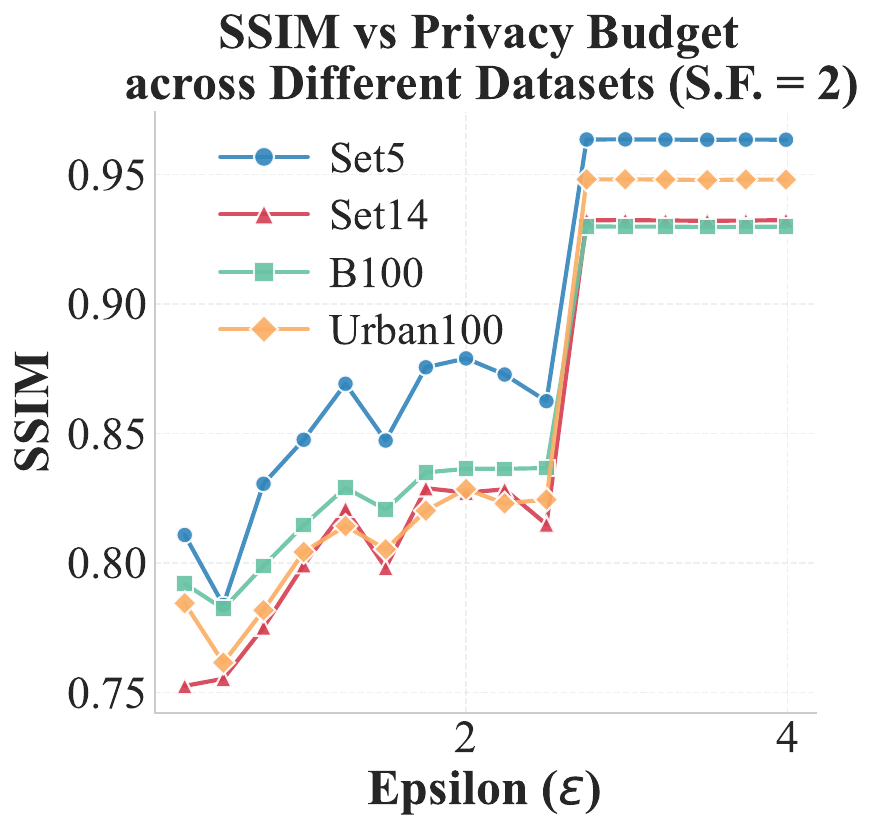}
        \caption{$\epsilon$ to SSIM, $\varsigma = 2$}
        \label{fig:sub4}
    \end{subfigure}
    
    % ==============
    \begin{subfigure}[b]{0.49\linewidth}
        \includegraphics[width=\linewidth]{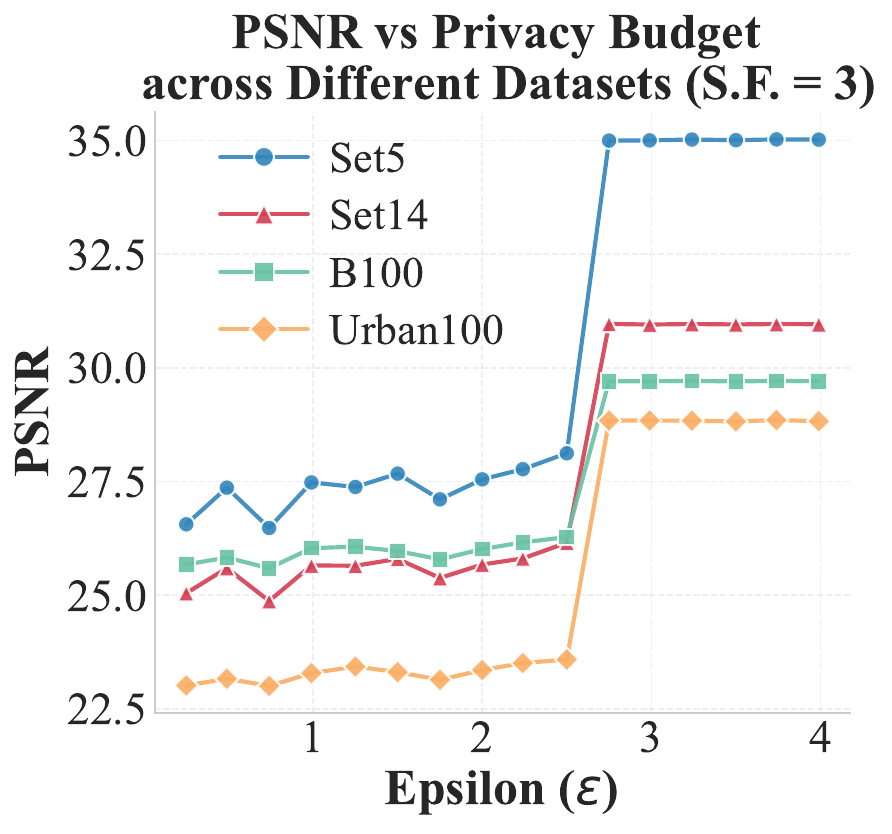}
        \caption{$\epsilon$ to PSNR, $\varsigma = 3$}
        \label{fig:sub2}
    \end{subfigure}
    \begin{subfigure}[b]{0.49\linewidth}
        \includegraphics[width=\linewidth]{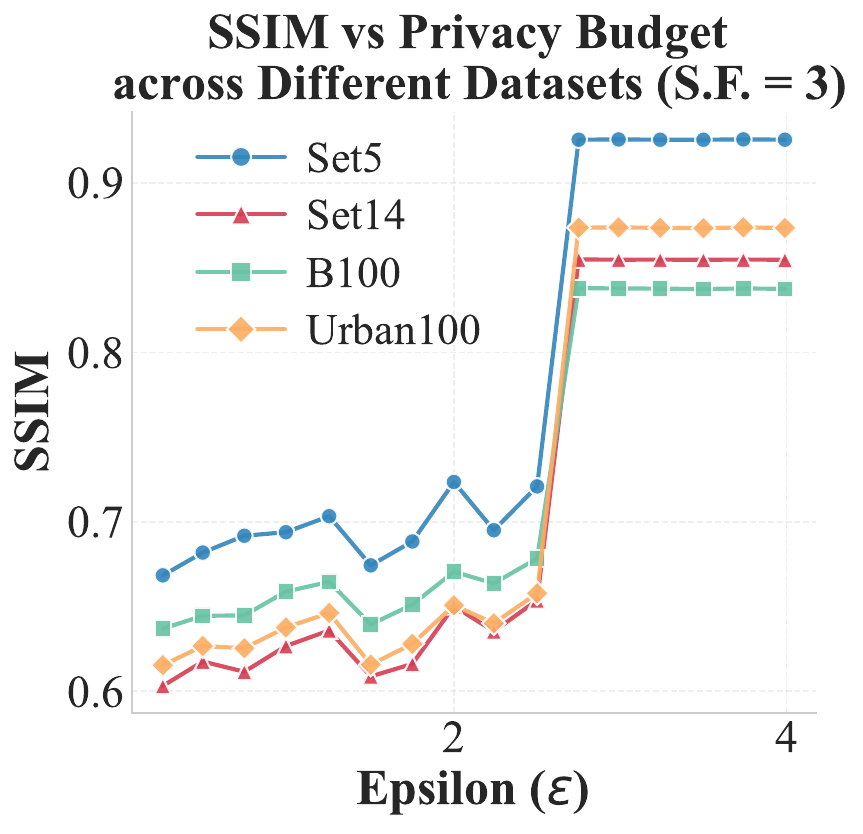}
        \caption{$\epsilon$ to SSIM, $\varsigma = 3$}
        \label{fig:sub5}
    \end{subfigure}
    
    % ==============
    
    \begin{subfigure}[b]{0.49\linewidth}
        \includegraphics[width=\linewidth]{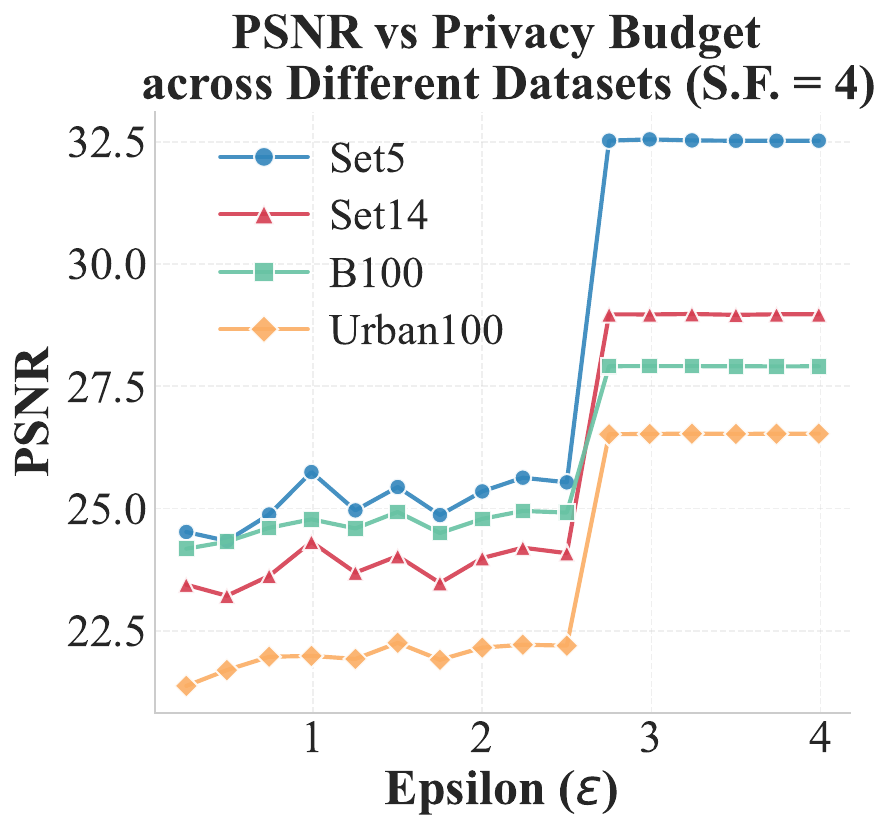}
        \caption{$\epsilon$ to PSNR, $\varsigma = 4$}
        \label{fig:sub3}
    \end{subfigure}
    \begin{subfigure}[b]{0.49\linewidth}
        \includegraphics[width=\linewidth]{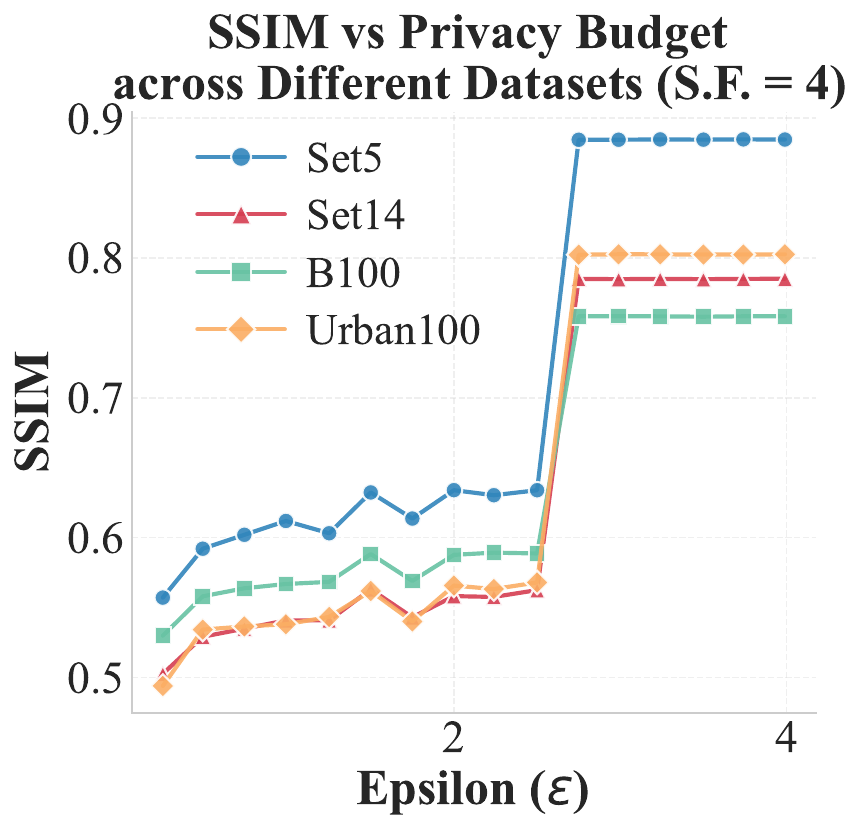}
        \caption{$\epsilon$ to SSIM, $\varsigma = 4$}
        \label{fig:sub6}
    \end{subfigure}

    \caption{\textit{Effect of privacy budgets on the overall PSNR and SSIM for different $\epsilon$ values.}}
    \label{fig:dp_epsilon}
\end{figure}

\paragraph{LDP + Baseline Performance.}
To evaluate privacy protection, we fine‑tune the pre‑trained FL‑RDSN under local DP.  
For ten clients, ten local epochs, and ten global rounds, we sweep the privacy budget \(\epsilon\in[0.25,4]\) using the \textsc{Opacus} \cite{opacus} library, follow the equation:
\begin{equation}
\small
\forall W\; \Pr\!\bigl[W(D)\bigr] \le \exp(\varepsilon)\,\Pr\!\bigl[W(D')\bigr] + \delta ,
\end{equation}
and report PSNR/SSIM in \cref{fig:dp_epsilon}.  
A broad plateau appears at \(\epsilon_{\text{best}}\!\in[2.5,3.0]\); we select \(\epsilon=2.75\) for all subsequent experiments and tabulate its effect in \cref{dp-comparison}.  Performance remains virtually unchanged, demonstrating that strong formal privacy can be added at negligible cost. Qualitative reconstructions in Appendix Sec. \ref{samples} exhibit high perceptual fidelity despite aggressive compression.

\paragraph{Watermark Effects.}
Table \ref{tab:wm_ablation} contrasts the vanilla FL-RDSN with our PPFL-RDSN$+\text{WM}$ on four DIV2K test sets: Set5, Set14, B100, and Urban100, scaling factor set to $4$.
We evaluate each model under four settings: (1) \emph{clean}; training, (2) \emph{fine-tuning} for 10 additional epochs (\textbf{+FT}), (3) \emph{70\% L1 pruning} of weights (\textbf{+Prune}), and (4) \emph{Gaussian weight noise} $\mathcal{N}(0,10^{-3})$ (\textbf{+Noise}).
Across all datasets and stress tests, the maximum \emph{PSNR} gap between the watermarked and vanilla models is below $0.08 \ dB$, while the largest $Top-1$ accuracy drop is less than 0.5\%. These results confirm that our watermarking scheme preserves predictive quality even under aggressive perturbations. 

% ========================== table 1 ============= 
\begin{table}[t]
\centering
\caption{\textit{Ablation study of watermarking.  
Super-resolution results are PSNR (dB) on DIV2K test subsets.
\textbf{+FT}: fine-tune 10 epochs;  
\textbf{+Prune}: 70\% L1 pruning;  
\textbf{+Noise}: add Gaussian weight noise $\mathcal{N}(0,10^{-3})$.  
Across all settings, the absolute difference between vanilla FL-RDSN and PPFL-RDSN+WM is $\le 0.08$ dB or 0.05\%.}}
\label{tab:wm_ablation}
\begin{tabular}{llcccc}
\toprule
\multicolumn{2}{c}{\textbf{Benchmark / Metric}} &
\textbf{Clean} & \textbf{+FT} & \textbf{+Prune} & \textbf{+Noise}\\
\midrule
\multicolumn{6}{c}{\emph{Super‑resolution – PSNR (dB)}}\\
\midrule
\multirow{2}{*}{Set5 ($\times$4)}   
 & Vanilla        & 32.69 & 32.67 & 32.10 & 32.65\\
 & PPFL\,+WM      & 32.52 & 32.50 & 31.92 & 32.48\\
\cmidrule{1-6}
\multirow{2}{*}{Set14 ($\times$4)}  
 & Vanilla        & 29.05 & 29.02 & 28.41 & 29.01\\
 & PPFL\,+WM      & 28.97 & 28.94 & 28.32 & 28.93\\
\cmidrule{1-6}
\multirow{2}{*}{B100 ($\times$4)}   
 & Vanilla        & 27.96 & 27.94 & 27.33 & 27.93\\
 & PPFL\,+WM      & 27.91 & 27.89 & 27.28 & 27.88\\
\cmidrule{1-6}
\multirow{2}{*}{Urban100 ($\times$4)}
 & Vanilla        & 26.72 & 26.69 & 26.05 & 26.70\\
 & PPFL\,+WM      & 26.52 & 26.50 & 25.84 & 26.51\\
\bottomrule
\end{tabular}
\end{table}
% ================== end =================

\begin{figure}[t!]
    \centering
    % 第一行：\varsigma = 2
    \begin{subfigure}[b]{0.49\linewidth}
        \centering
        \includegraphics[width=\linewidth, height=4cm]{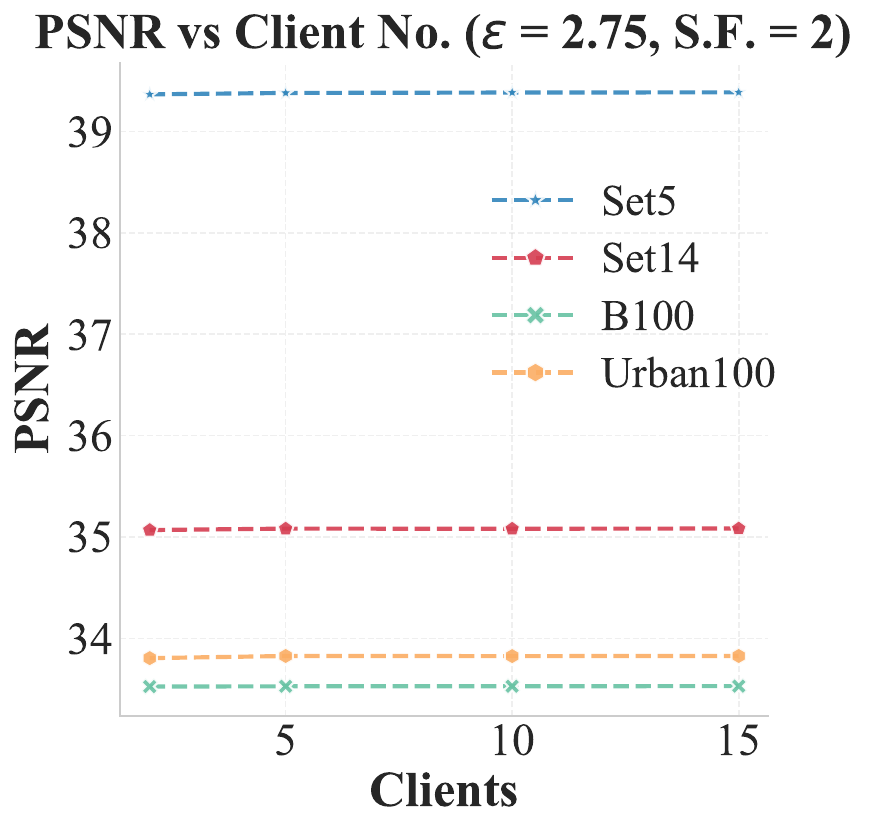}
        \caption{\(client\) to PSNR, \(\varsigma = 2\)}
        \label{fig:sub1}
    \end{subfigure}%
    \begin{subfigure}[b]{0.49\linewidth}
        \centering
        \includegraphics[width=\linewidth, height=4cm]{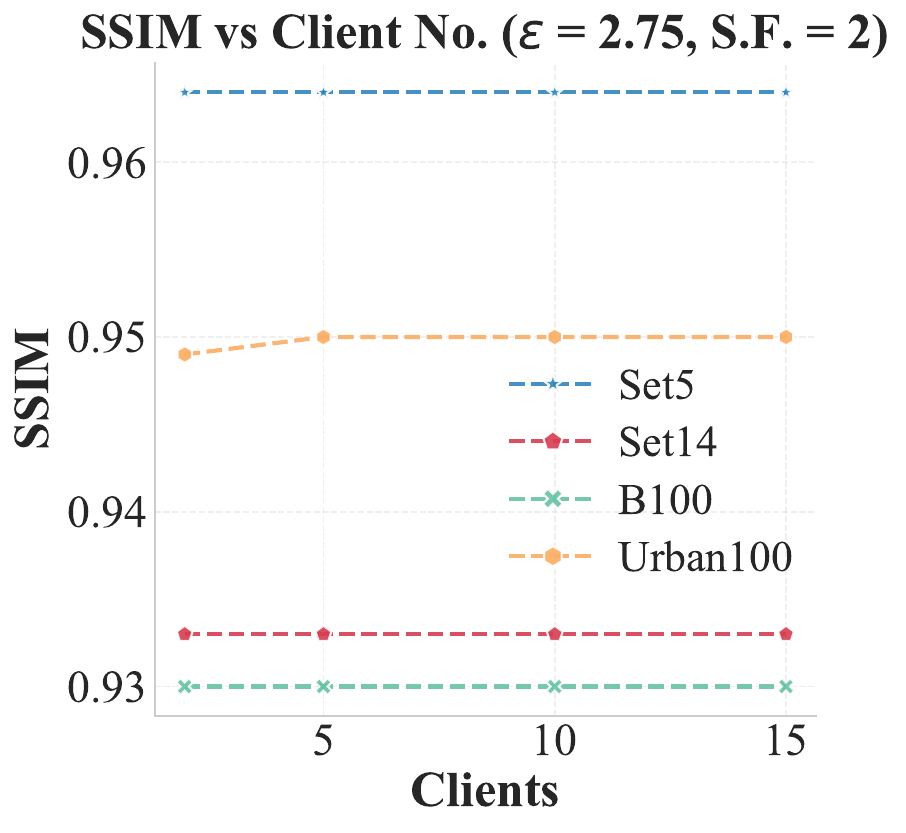}
        \caption{\(client\) to SSIM, \(\varsigma = 2\)}
        \label{fig:sub4}
    \end{subfigure}
    
    % 第二行：\varsigma = 3
    \begin{subfigure}[b]{0.49\linewidth}
        \centering
        \includegraphics[width=\linewidth, height=4cm]{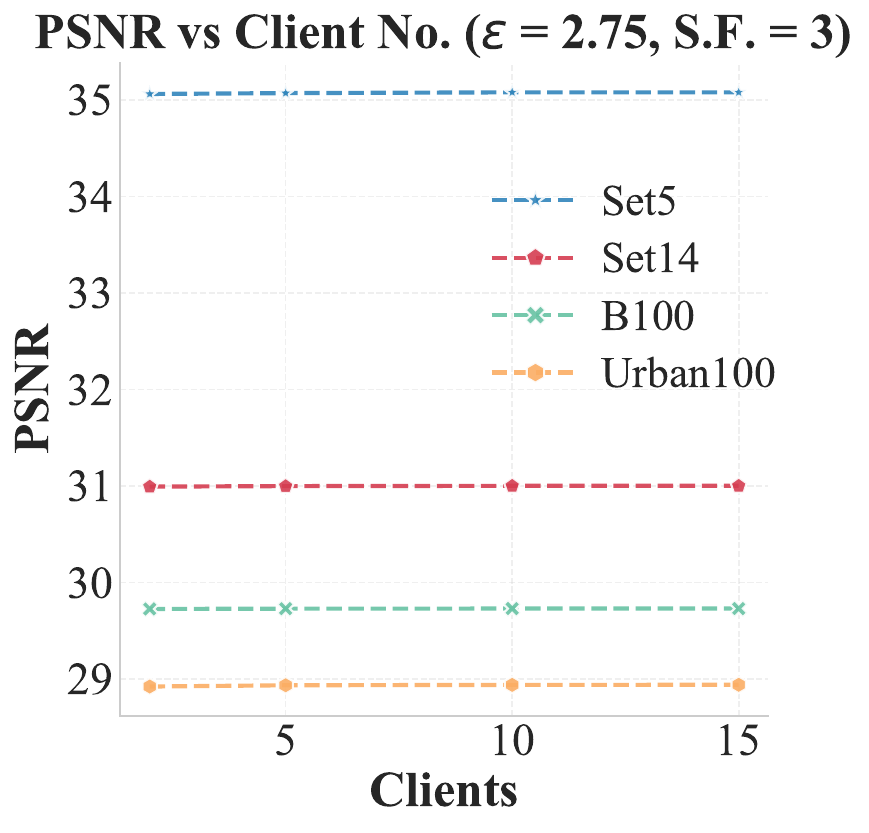}
        \caption{\(client\) to PSNR, \(\varsigma = 3\)}
        \label{fig:sub2}
    \end{subfigure}%
    \begin{subfigure}[b]{0.49\linewidth}
        \centering
        \includegraphics[width=\linewidth, height=4cm]{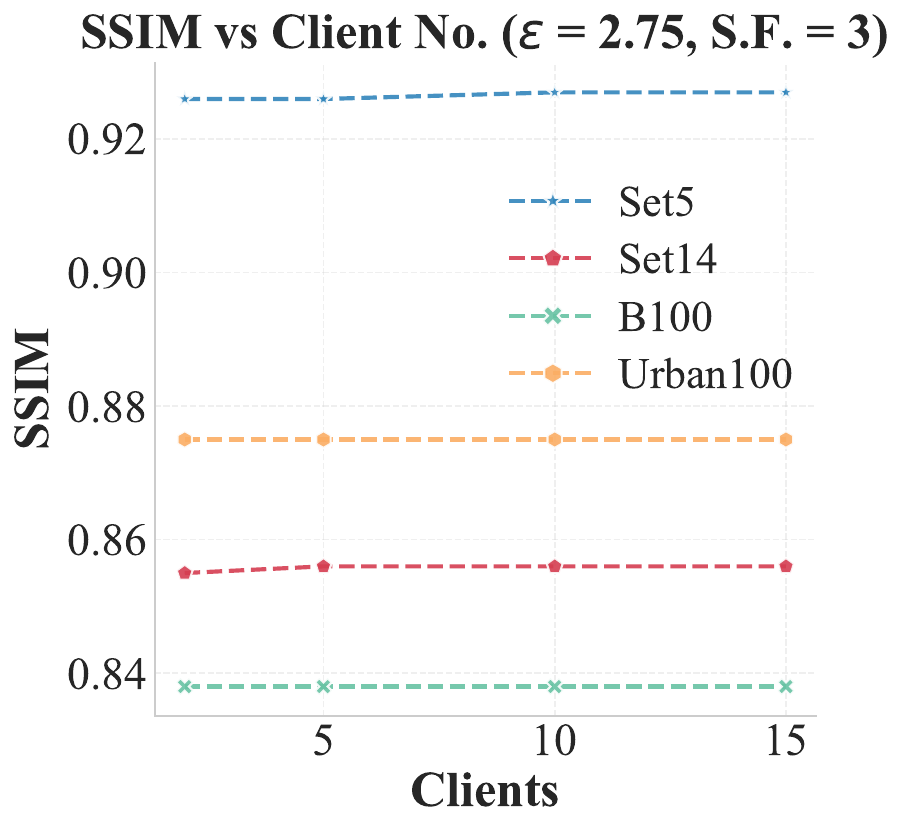}
        \caption{\(client\) to SSIM, \(\varsigma = 3\)}
        \label{fig:sub5}
    \end{subfigure}
    
    % 第三行：\varsigma = 4
    \begin{subfigure}[b]{0.49\linewidth}
        \centering
        \includegraphics[width=\linewidth, height=4cm]{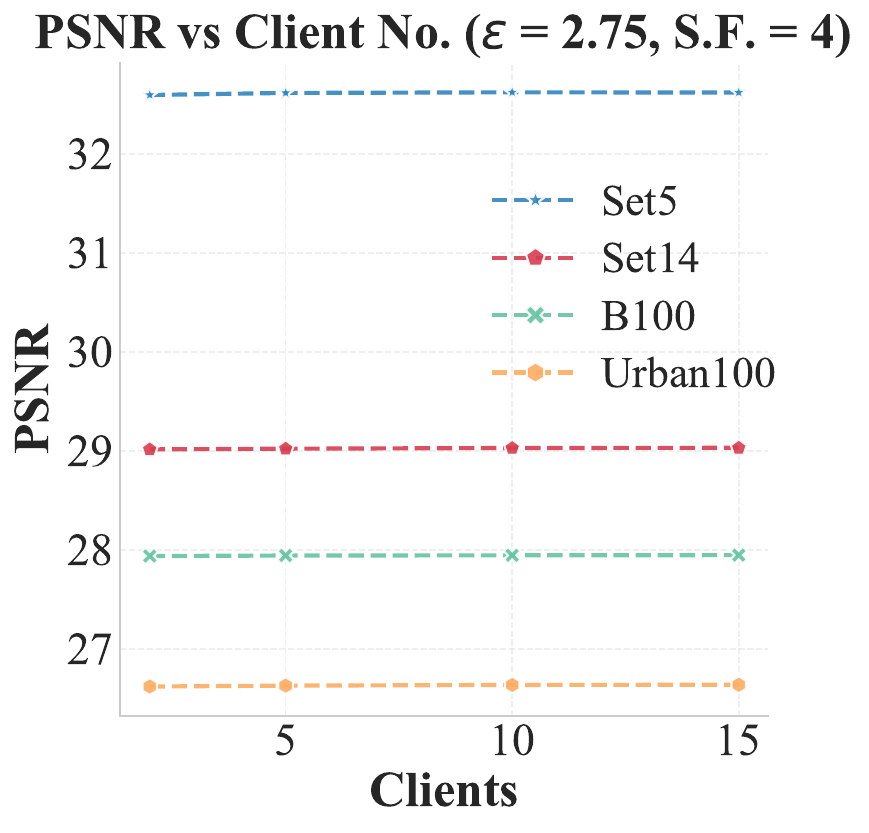}
        \caption{\(client\) to PSNR, \(\varsigma = 4\)}
        \label{fig:sub3}
    \end{subfigure}%
    \begin{subfigure}[b]{0.49\linewidth}
        \centering
        \includegraphics[width=\linewidth, height=4cm]{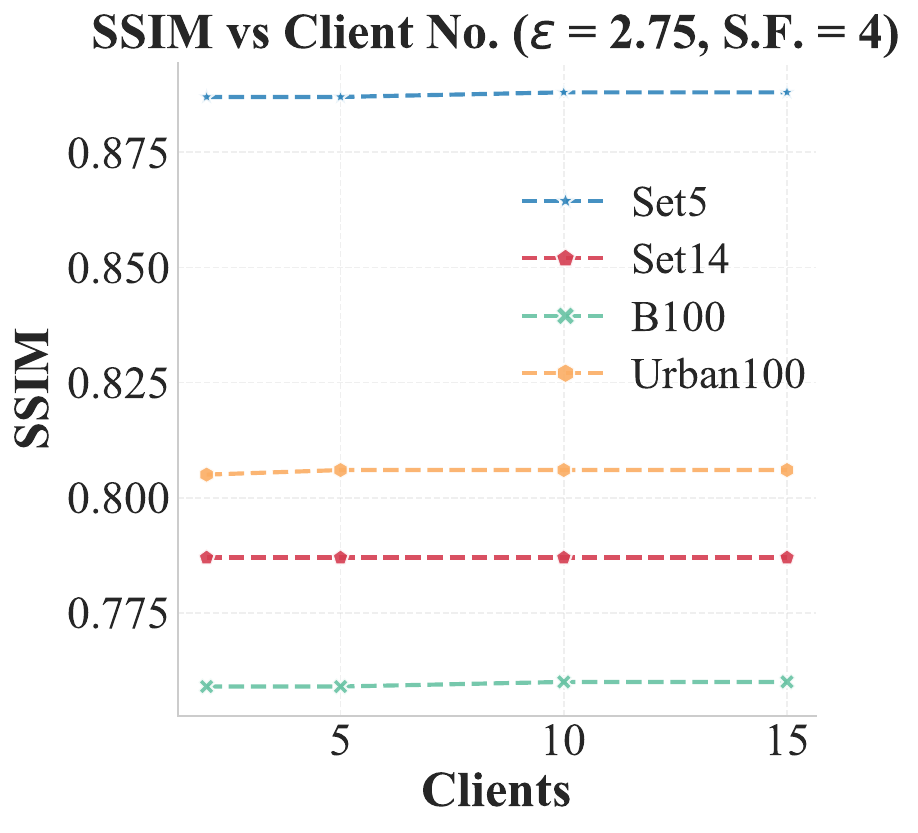}
        \caption{\(client\) to SSIM, \(\varsigma = 4\)}
        \label{fig:sub6}
    \end{subfigure}

    \caption{\textit{Effect of privacy budgets on the overall PSNR and SSIM for different \(\epsilon\) values.}}
    \label{fig:dp_client}
\end{figure}

\paragraph{Scalability with Client Pool Size.}
\cref{fig:dp_client} plots PSNR/SSIM as the number of clients grows.  Even with DP enabled, quality degrades only marginally, confirming that our framework scales gracefully and benefits from the added computational capacity of larger federations.

\begin{figure}[t!]
    \centering
    % \varsigma = 2
    \begin{subfigure}[b]{0.49\linewidth}
        \centering
        \includegraphics[width=\linewidth, height=4cm]{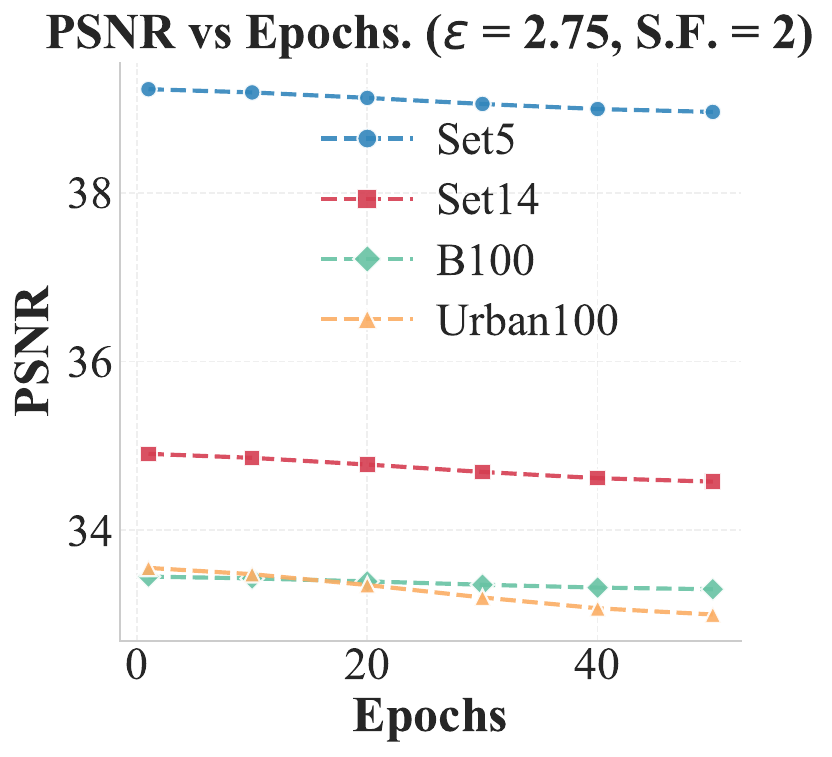}
        \caption{\(epoch\) to PSNR, \(\varsigma = 2\)}
        \label{fig:epoch_sub1}
    \end{subfigure}%
    \begin{subfigure}[b]{0.49\linewidth}
        \centering
        \includegraphics[width=\linewidth, height=4cm]{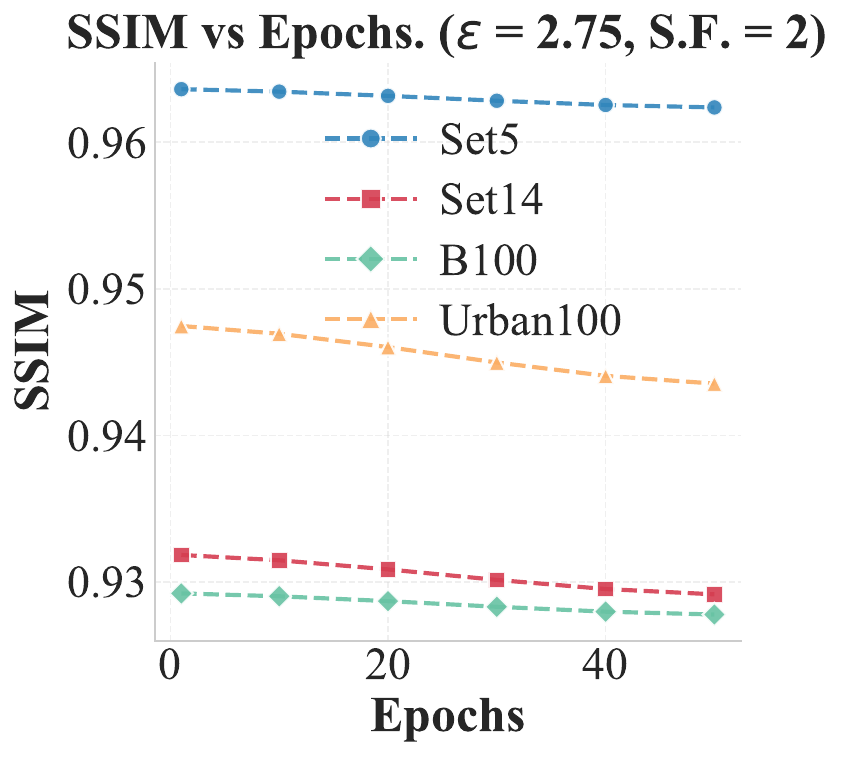}
        \caption{\(epoch\) to SSIM, \(\varsigma = 2\)}
        \label{fig:epoch_sub4}
    \end{subfigure}

    % \varsigma = 3
    \begin{subfigure}[b]{0.49\linewidth}
        \centering
        \includegraphics[width=\linewidth, height=4cm]{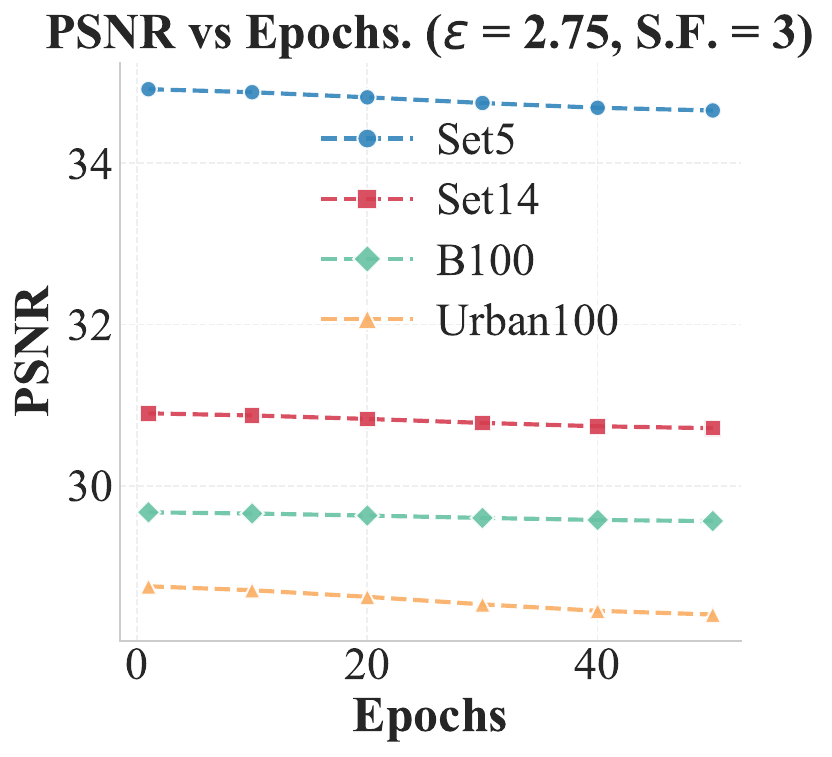}
        \caption{\(epoch\) to PSNR, \(\varsigma = 3\)}
        \label{fig:epoch_sub2}
    \end{subfigure}%
    \begin{subfigure}[b]{0.49\linewidth}
        \centering
        \includegraphics[width=\linewidth, height=4cm]{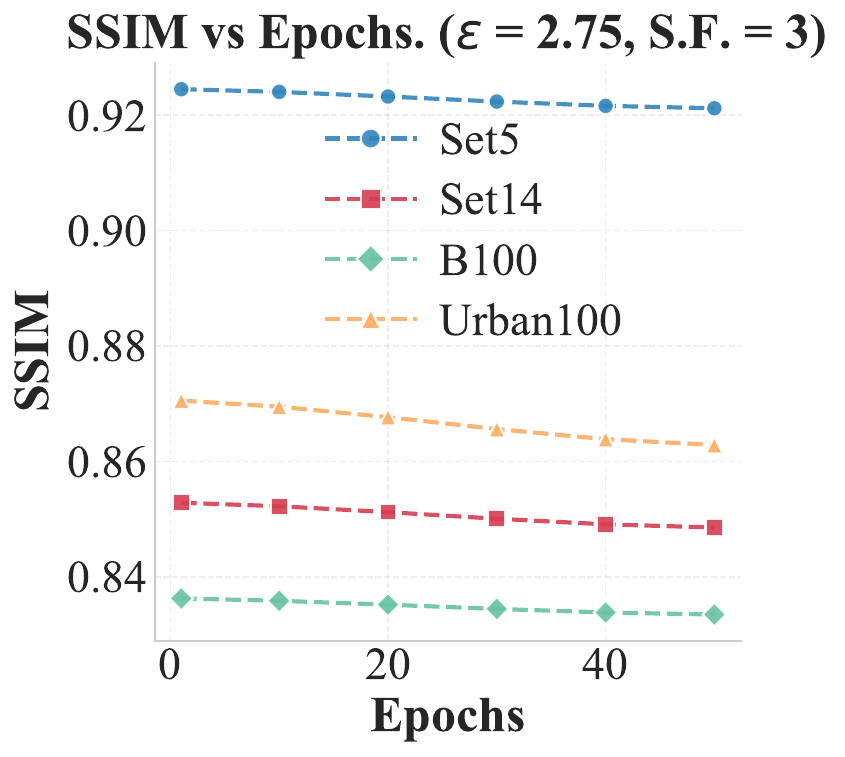}
        \caption{\(epoch\) to SSIM, \(\varsigma = 3\)}
        \label{fig:epoch_sub5}
    \end{subfigure}

    % \varsigma = 4
    \begin{subfigure}[b]{0.49\linewidth}
        \centering
        \includegraphics[width=\linewidth, height=4cm]{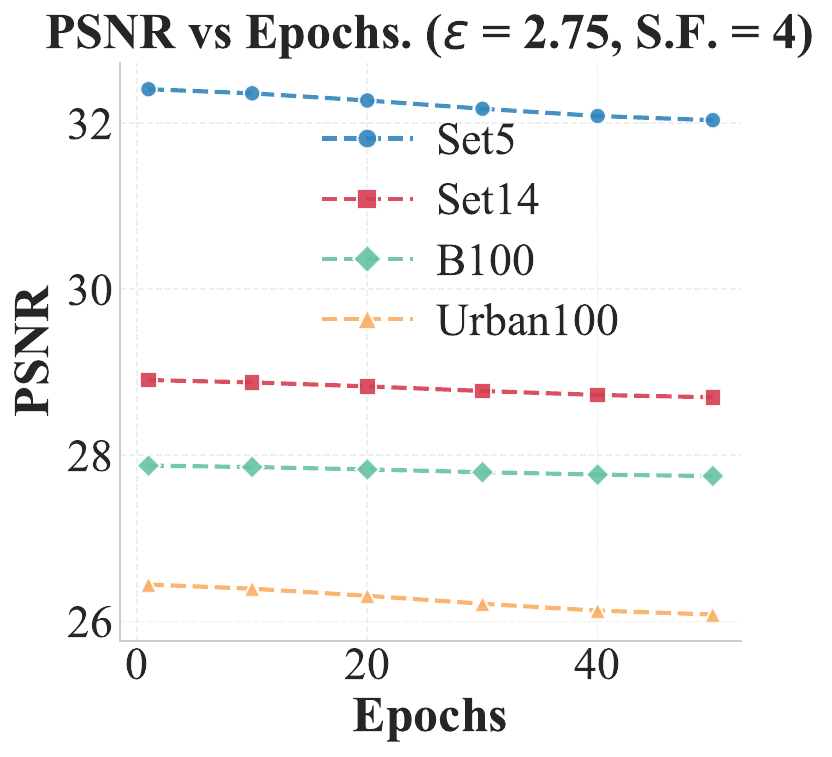}
        \caption{\(epoch\) to PSNR, \(\varsigma = 4\)}
        \label{fig:epoch_sub3}
    \end{subfigure}%
    \begin{subfigure}[b]{0.49\linewidth}
        \centering
        \includegraphics[width=\linewidth, height=4cm]{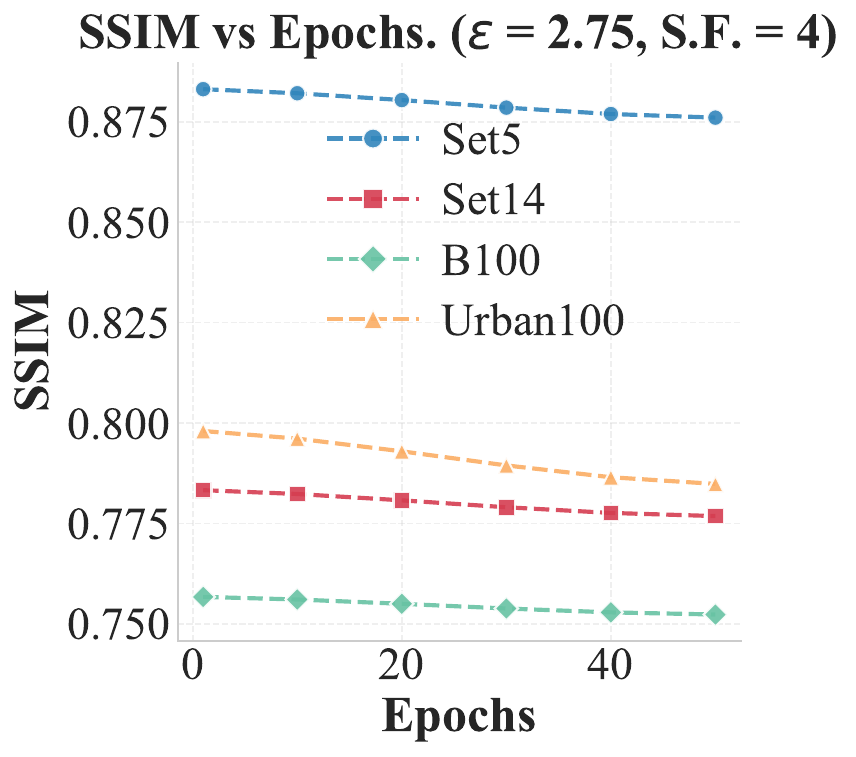}
        \caption{\(epoch\) to SSIM, \(\varsigma = 4\)}
        \label{fig:epoch_sub6}
    \end{subfigure}

    \caption{\textit{Effect of the number of epochs on the overall PSNR and SSIM for different \(\epsilon\) values.}}
    \label{fig:epoch-dp}
\end{figure}

\paragraph{Effect of Local Epochs.}
\cref{fig:epoch-dp} explores longer local training schedules.  
Under DP, very large numbers of local epochs introduce a modest quality drop; therefore, for best overall throughput, one should jointly tune the \emph{client count} and \emph{local epochs} to stay on the flat part of the curve.

% =============client dp===============
% =================end dp=====================

% ==================epoch============

% ==================end epoch============
% \input{related}

\section{Conclusion and Discussion}
\label{conclusion}
In this paper, we have presented an innovative Privacy-Preserving Federated Learning (PPFL) framework for training Residual Dense Spatial Networks (RDSN) and designed a multi-party Encryption-then-Lossy-Compression scheme based on \cite{zhang2010lossy}. Our experimental results show the proposed PPFL-RDSN framework's advantages in maintaining high-quality and robust lossy image reconstruction while enhancing privacy and security. The clients perform their local differentially private training to obtain the initial model. After aggregation and model integrity verification, they execute the multi-party EtLC scheme and determine whether the initial model demands refinement iteratively. We show that the PPFL-RDSN training system is scalable to a larger client pool and provides strong privacy guarantees offered by LDP and protection against model poisoning or substitution by untrusted aggregators. Our framework makes an important contribution in employing privacy-preserving federated learning for image processing tasks, paving the way for more privacy-protecting, secure, and efficient collaborative learning solutions essential for data-intensive applications with distributed datasets. Future research directions can focus on optimizing the RDSN structures to improve the model performance further, incorporating other privacy protection mechanisms to address a broader set of privacy and security threats to enhance the framework. Also, exploring hybrid approaches that combine multiple privacy techniques, such as DP and multiparty computation, is a potentially promising research direction to address practical challenges for real-world applications.

\textbf{Limitations.}
Our design emphasizes a practical baseline—high-frequency DCT perturbation with fixed $(\tau, C_{\max})$ and moments-accountant composition, a sparse-index watermark with standard key management, and an honest-but-curious multi-party EtLC prototype with FedAvg; adaptive frequency protection, richer collusion models, and secure or Byzantine-robust aggregation are natural extensions. Empirically, we evaluate on canonical SR datasets with a 10-client simulation and representative $(\epsilon,\delta)$; scaling to larger and heterogeneous client pools and wide-area settings, broadening to non-vision tasks, and expanding robustness and $\epsilon$–utility sweeps are promising directions.

\section*{Acknowledgments}
This research is partially supported by the University of Pittsburgh Center for Research Computing through the resources provided. Specifically, this work used the HTC cluster, which is supported by NIH award number S10OD028483. 
% The authors would appreciate the open-source contributions of Wang et al. \cite{wang2022novel} and Zhang et al. \cite{zhang2023pfllib}. Meanwhile, the authors would appreciate the anonymous reviewers and the editor for their precious comments and suggestions.
\bibliographystyle{IEEEtran}
% \bibliography{ref}

%

\newpage
\appendix

% ===============================

\subsection{Comparisons}
\label{comparison}
In this section, we draw a comparison between our model performance and the state-of-the-art model \cite{wang2022novel}. We use PSNR and SSIM metrics as the evaluation criterion, using 4 testing datasets: Set5, Set14, BSDB100, and Urban100, and also have 3 scaling factors: 2, 3, and 4. Under the same environment settings, same training epochs, same computing powers, etc., with greater averages of PNSR or SSIM values, the model performance is considered better.

% 多图并排展示
\begin{figure*}[t!]
\centering
% 第一行子图
\begin{subfigure}{0.32\textwidth}
\includegraphics[width=\textwidth,height=5cm]{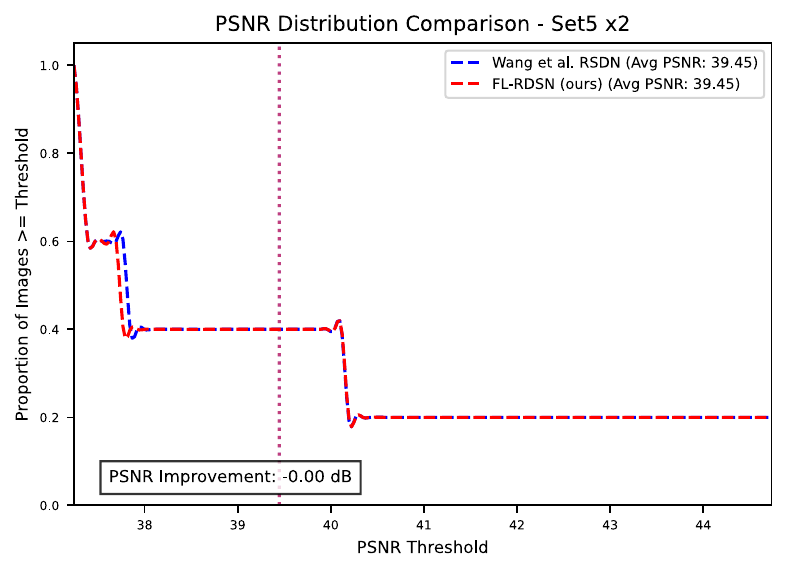}
\caption{Set5, PSNR, \(\varsigma = 2\)}
\label{fig:set5-2}
\end{subfigure}
\hfill
\begin{subfigure}{0.32\textwidth}
\includegraphics[width=\textwidth,height=5cm]{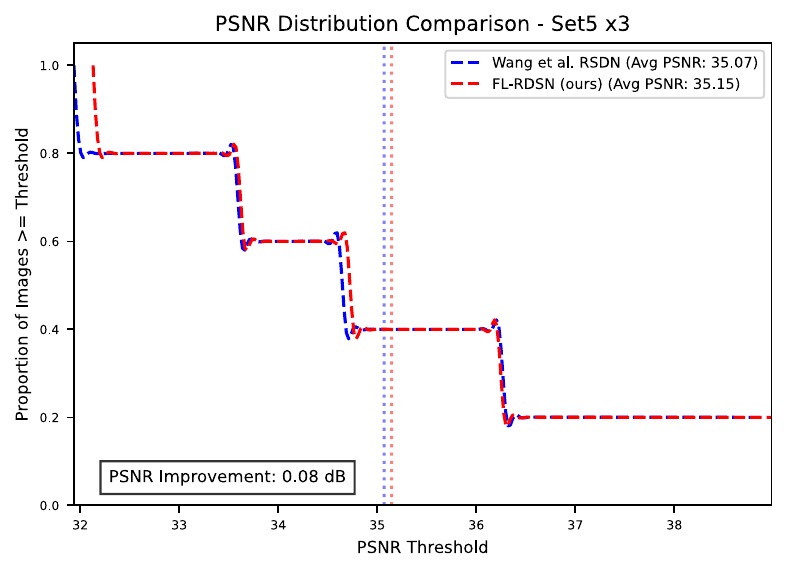}
\caption{Set5, PSNR, \(\varsigma = 3\)}
\label{fig:set5-3}
\end{subfigure}
\hfill
\begin{subfigure}{0.32\textwidth}
\includegraphics[width=\textwidth,height=5cm]{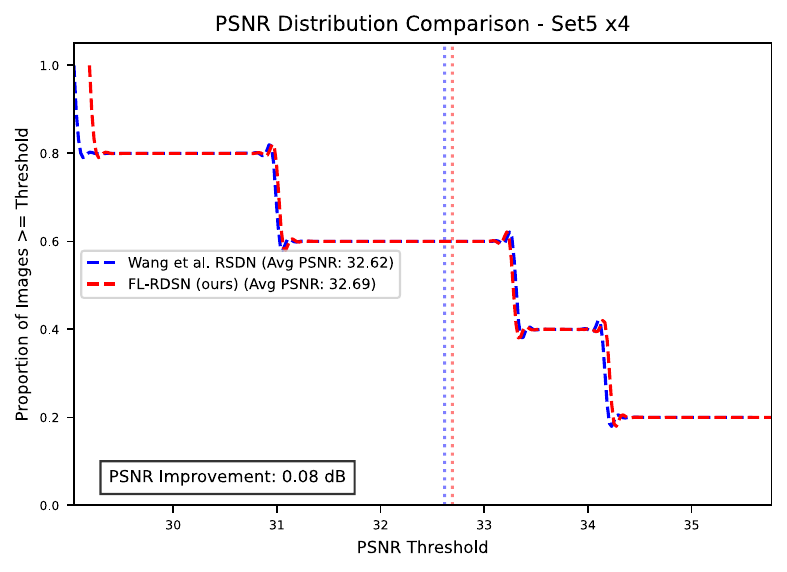}
\caption{Set5, PSNR, \(\varsigma = 4\)}
\label{fig:set5-4}
\end{subfigure}

% 第二行子图
\begin{subfigure}{0.32\textwidth}
\includegraphics[width=\textwidth,height=5cm]{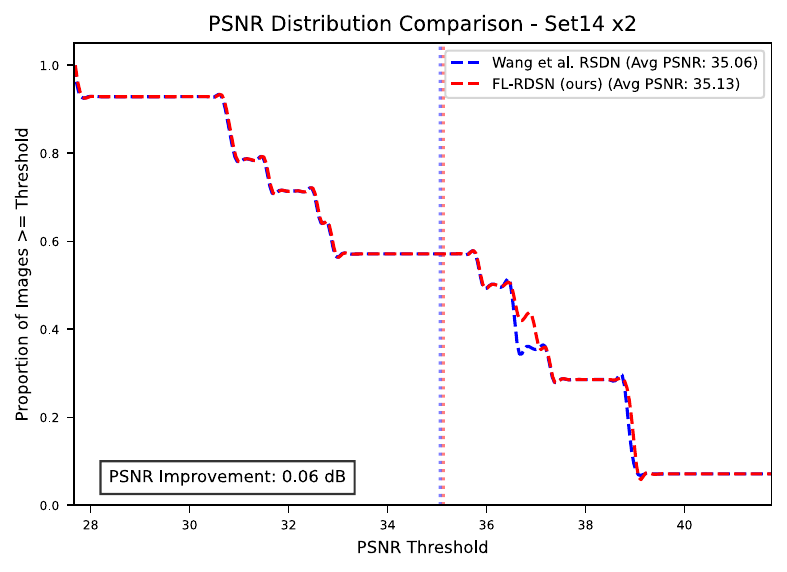}
\caption{Set14, PSNR, \(\varsigma = 2\)}
\label{fig:set14-2}
\end{subfigure}
\hfill
\begin{subfigure}{0.32\textwidth}
\includegraphics[width=\textwidth,height=5cm]{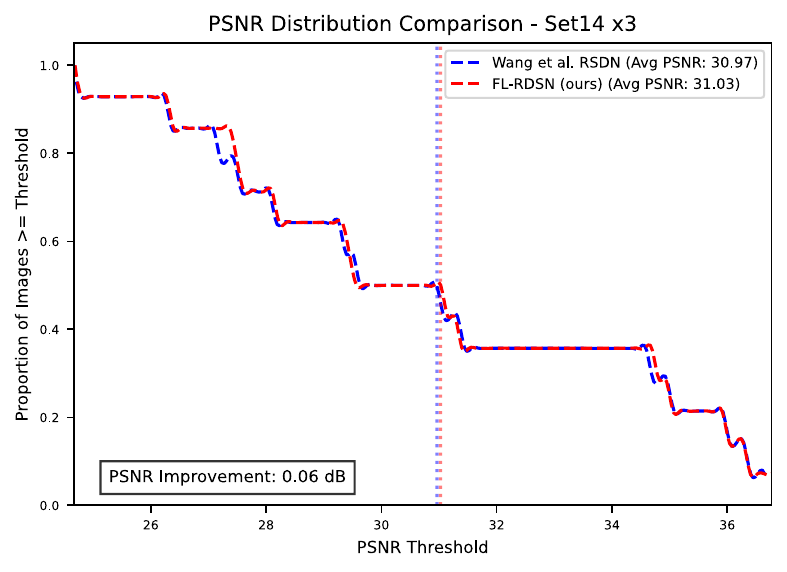}
\caption{Set14, PSNR, \(\varsigma = 3\)}
\label{fig:set14-3}
\end{subfigure}
\hfill
\begin{subfigure}{0.32\textwidth}
\includegraphics[width=\textwidth,height=5cm]{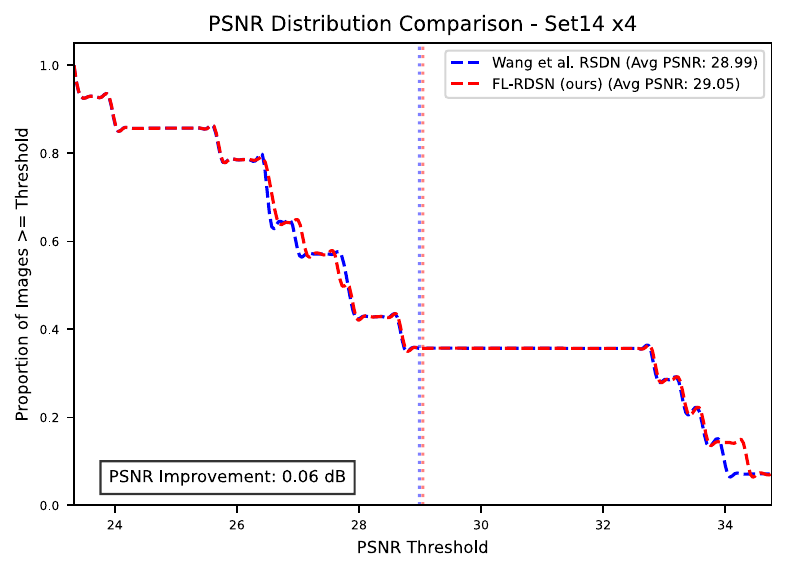}
\caption{Set14, PSNR, \(\varsigma = 4\)}
\label{fig:set14-4}
\end{subfigure}

% 第三行子图
\begin{subfigure}{0.32\textwidth}
\includegraphics[width=\textwidth,height=5cm]{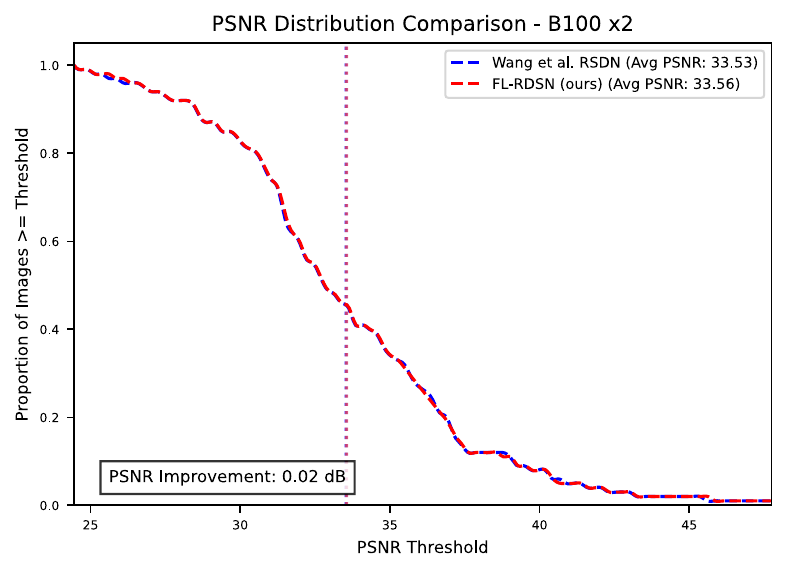}
\caption{BSDB100, PSNR, \(\varsigma = 2\)}
\label{fig:b100-2}
\end{subfigure}
\hfill
\begin{subfigure}{0.32\textwidth}
\includegraphics[width=\textwidth,height=5cm]{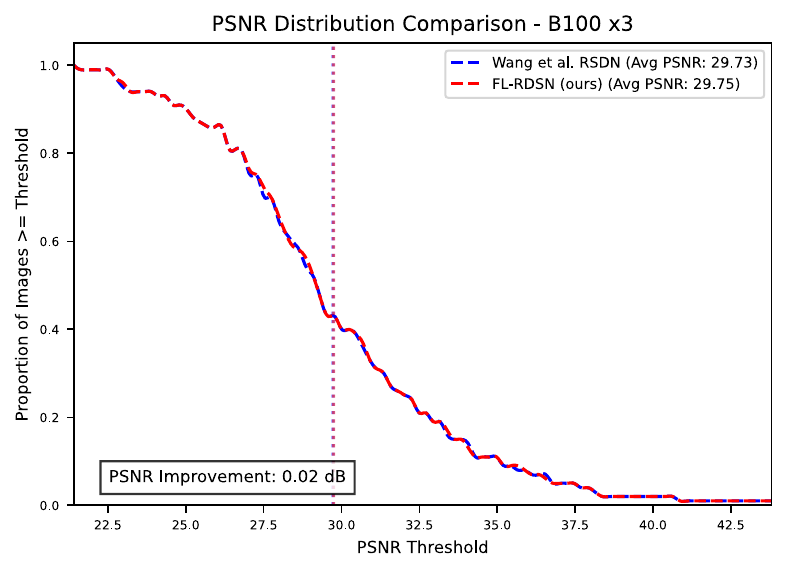}
\caption{BSDB100, PSNR, \(\varsigma = 3\)}
\label{fig:b100-3}
\end{subfigure}
\hfill
\begin{subfigure}{0.32\textwidth}
\includegraphics[width=\textwidth,height=5cm]{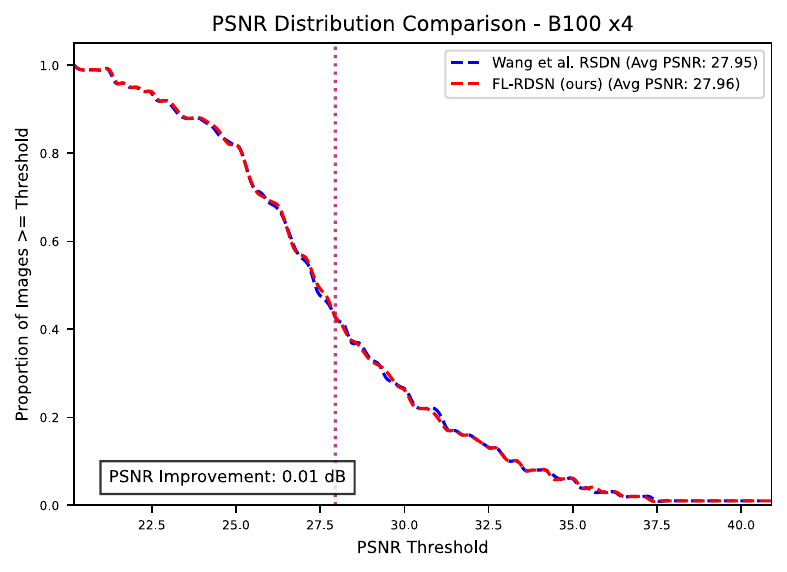}
\caption{BSDB100, PSNR, \(\varsigma = 4\)}
\label{fig:b100-4}
\end{subfigure}

% 第四行子图
\begin{subfigure}{0.32\textwidth}
\includegraphics[width=\textwidth,height=5cm]{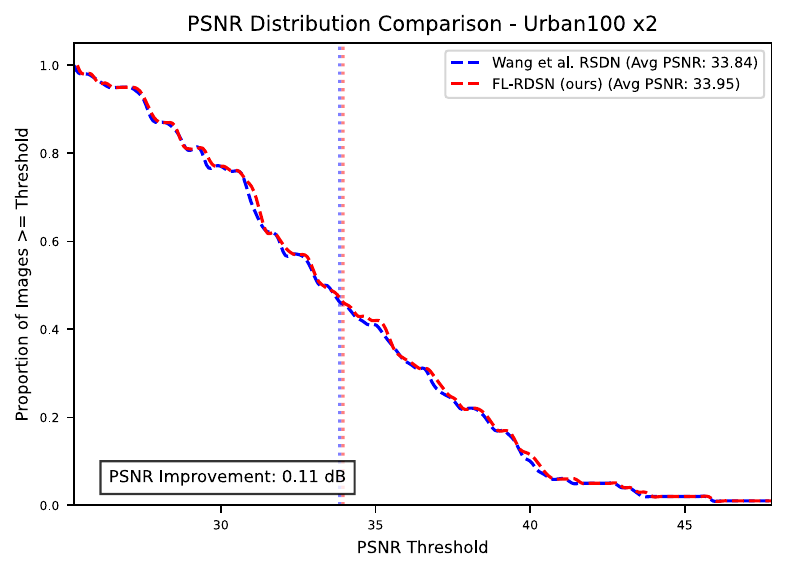}
\caption{Urban100, PSNR, \(\varsigma = 2\)}
\label{fig:urban100-2}
\end{subfigure}
\hfill
\begin{subfigure}{0.32\textwidth}
\includegraphics[width=\textwidth,height=5cm]{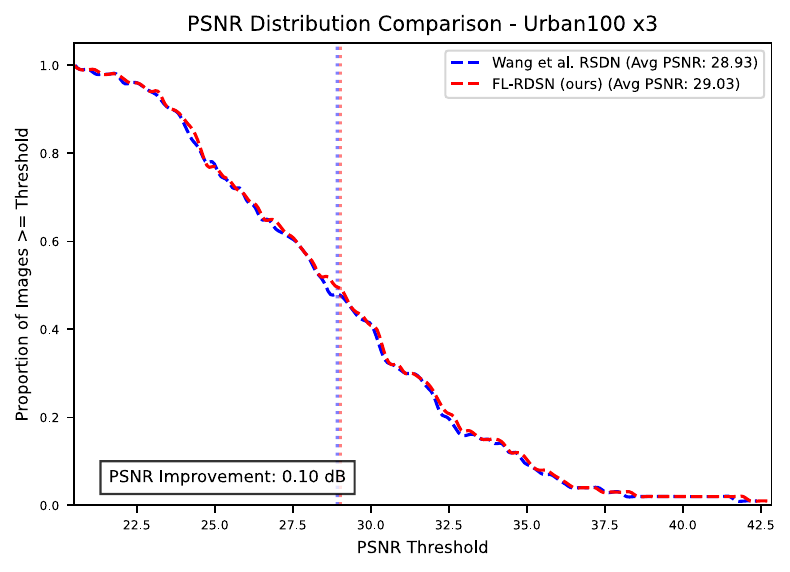}
\caption{Urban100, PSNR, \(\varsigma = 3\)}
\label{fig:urban100-3}
\end{subfigure}
\hfill
\begin{subfigure}{0.32\textwidth}
\includegraphics[width=\textwidth,height=5cm]{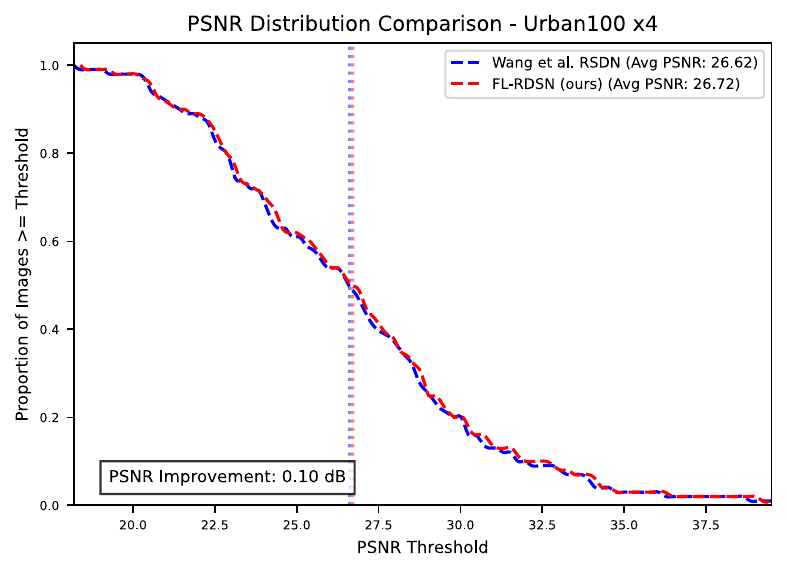}
\caption{Urban100, PSNR, \(\varsigma = 4\)}
\label{fig:urban100-4}
\end{subfigure}

\caption{Performance comparison for PSNR between State-of-the-Art and our FL-RDSN}
\label{fig:psnr-comparison}
\end{figure*}

% 多图并排展示
\begin{figure*}[htb!]
\centering
% 第一行子图
\begin{subfigure}{0.32\textwidth}
\includegraphics[width=\textwidth,height=5cm]{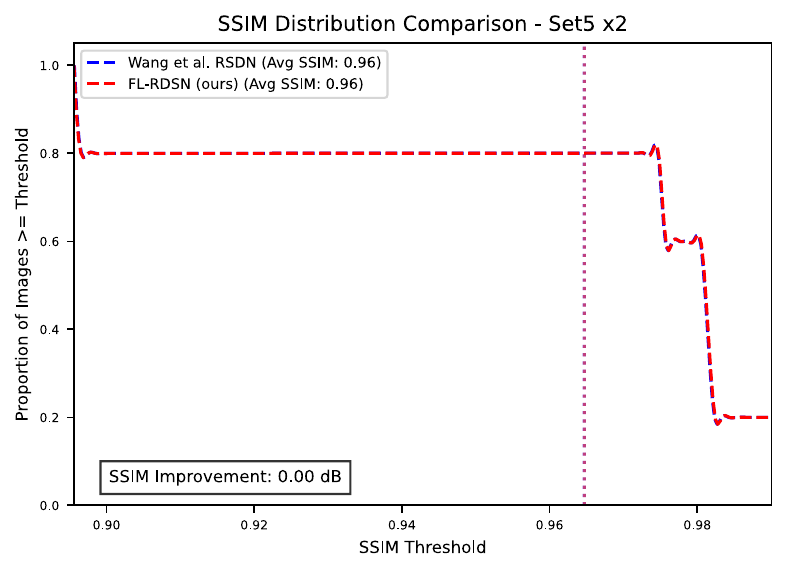}
\caption{Set5, SSIM, \(\varsigma = 2\)}
\label{fig:set5-2-ssim}
\end{subfigure}
\hfill
\begin{subfigure}{0.32\textwidth}
\includegraphics[width=\textwidth,height=5cm]{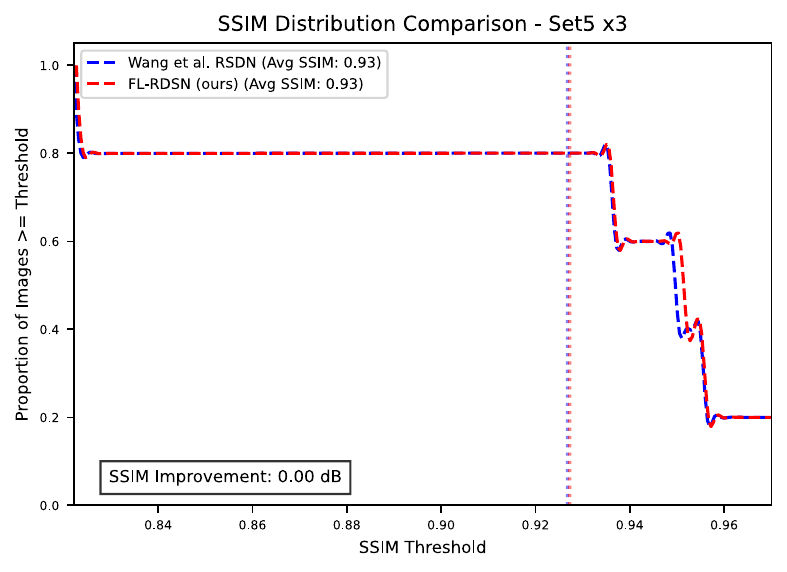}
\caption{Set5, SSIM, \(\varsigma = 3\)}
\label{fig:set5-3-ssim}
\end{subfigure}
\hfill
\begin{subfigure}{0.32\textwidth}
\includegraphics[width=\textwidth,height=5cm]{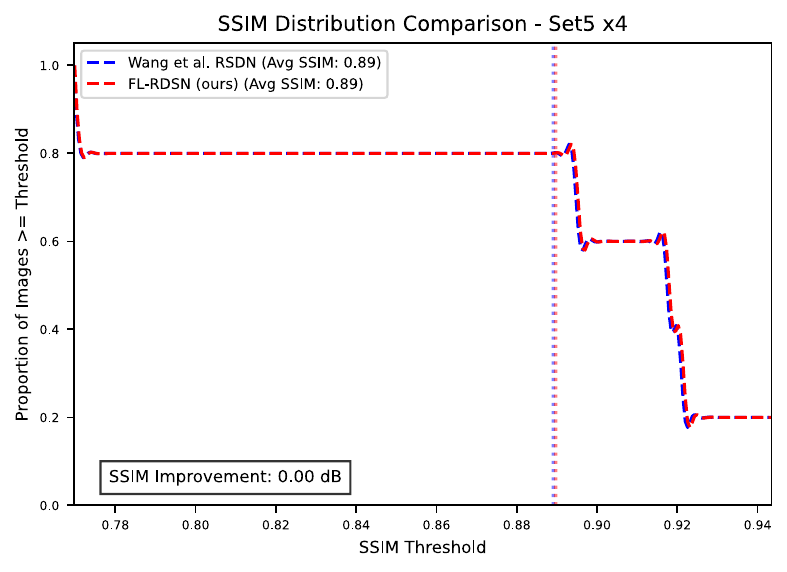}
\caption{Set5, SSIM, \(\varsigma = 4\)}
\label{fig:set5-4-ssim}
\end{subfigure}

% 第二行子图
\begin{subfigure}{0.32\textwidth}
\includegraphics[width=\textwidth,height=5cm]{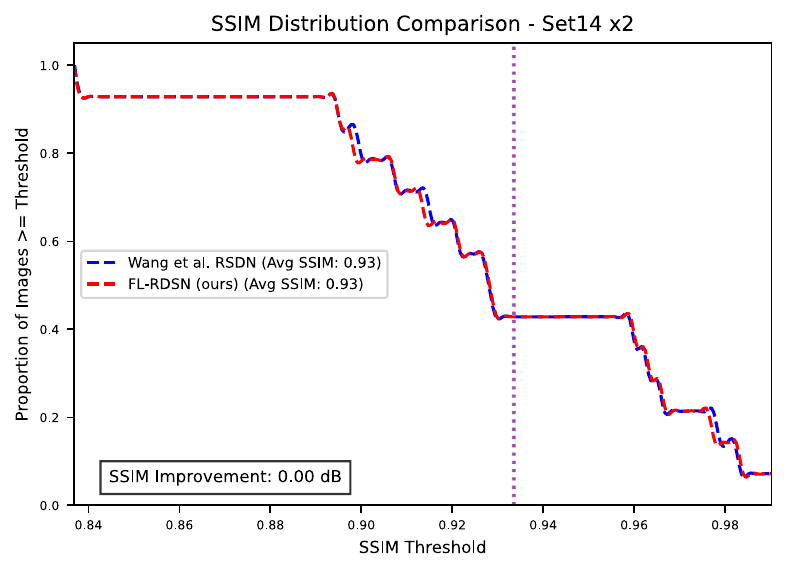}
\caption{Set14, SSIM, \(\varsigma = 2\)}
\label{fig:set14-2-ssim}
\end{subfigure}
\hfill
\begin{subfigure}{0.32\textwidth}
\includegraphics[width=\textwidth,height=5cm]{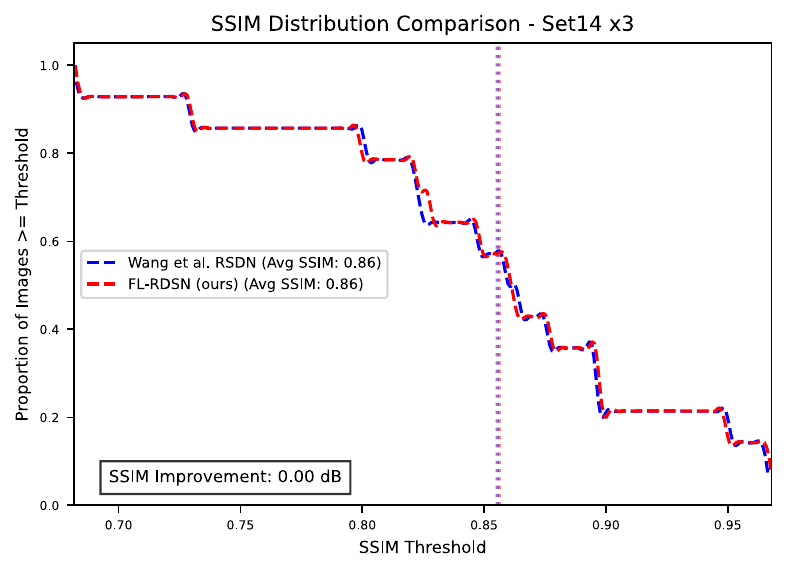}
\caption{Set14, SSIM, \(\varsigma = 3\)}
\label{fig:set14-3-ssim}
\end{subfigure}
\hfill
\begin{subfigure}{0.32\textwidth}
\includegraphics[width=\textwidth,height=5cm]{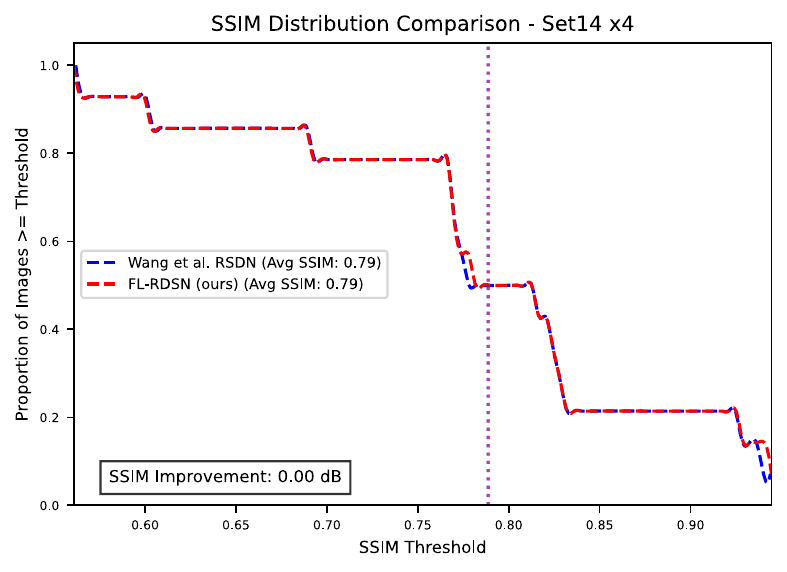}
\caption{Set14, SSIM, \(\varsigma = 4\)}
\label{fig:set14-4-ssim}
\end{subfigure}

% 第三行子图
\begin{subfigure}{0.32\textwidth}
\includegraphics[width=\textwidth,height=5cm]{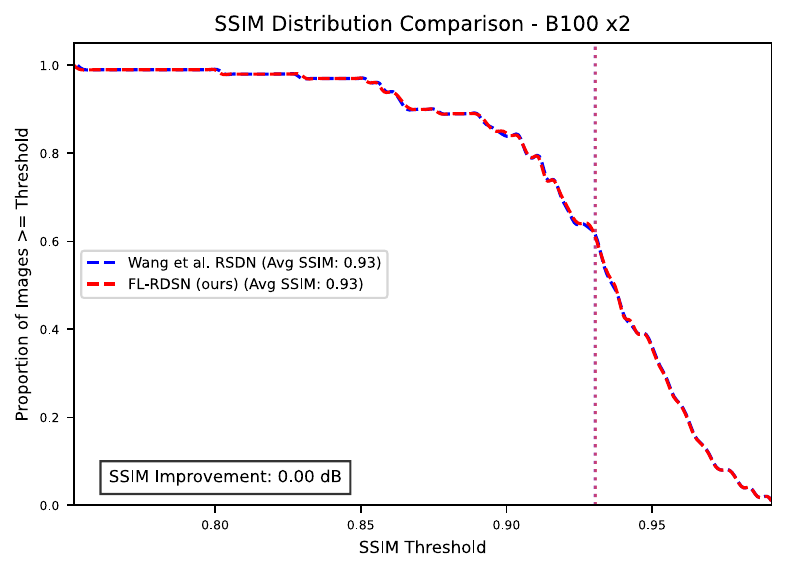}
\caption{BSDB100, SSIM, \(\varsigma = 2\)}
\label{fig:b100-2-ssim}
\end{subfigure}
\hfill
\begin{subfigure}{0.32\textwidth}
\includegraphics[width=\textwidth,height=5cm]{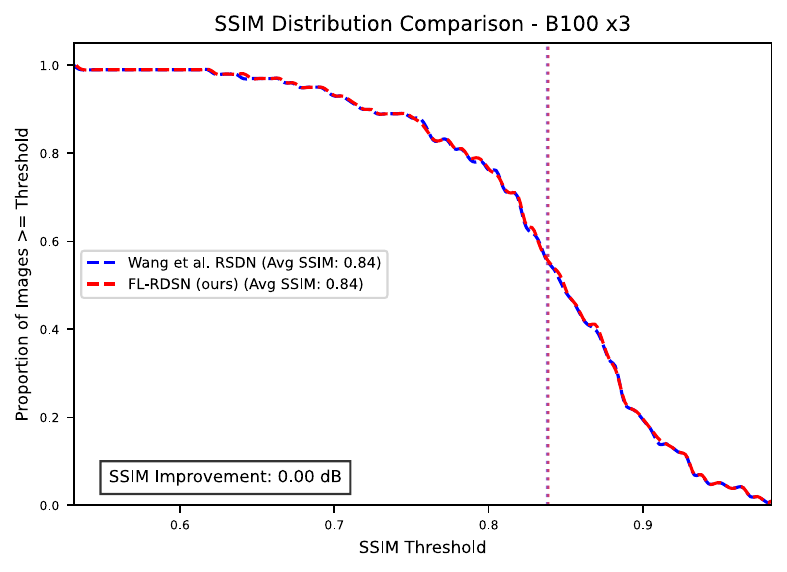}
\caption{BSDB100, SSIM, \(\varsigma = 3\)}
\label{fig:b100-3-ssim}
\end{subfigure}
\hfill
\begin{subfigure}{0.32\textwidth}
\includegraphics[width=\textwidth,height=5cm]{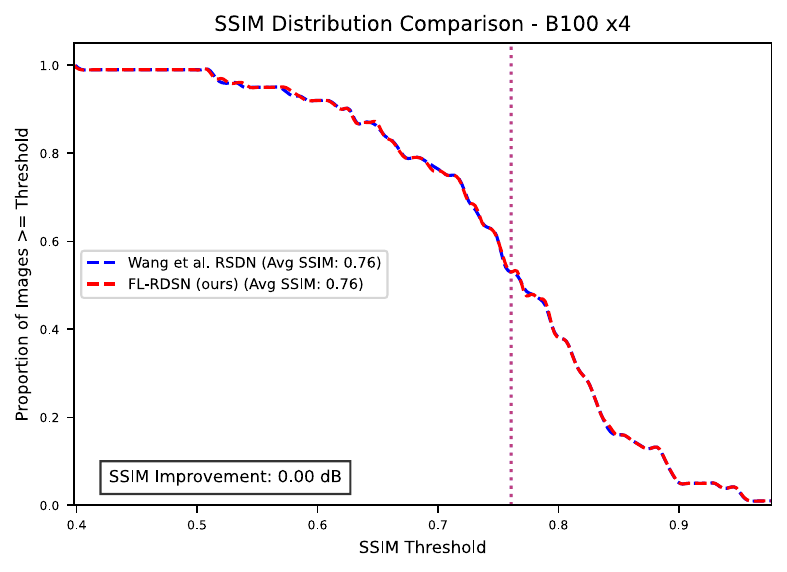}
\caption{BSDB100, SSIM, \(\varsigma = 4\)}
\label{fig:b100-4-ssim}
\end{subfigure}

% 第四行子图
\begin{subfigure}{0.32\textwidth}
\includegraphics[width=\textwidth,height=5cm]{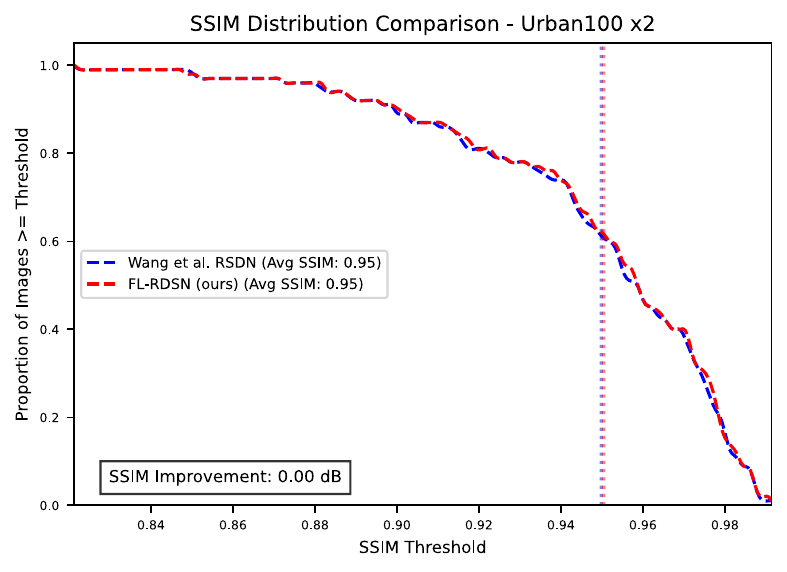}
\caption{Urban100, SSIM, \(\varsigma = 2\)}
\label{fig:urban100-2-ssim}
\end{subfigure}
\hfill
\begin{subfigure}{0.32\textwidth}
\includegraphics[width=\textwidth,height=5cm]{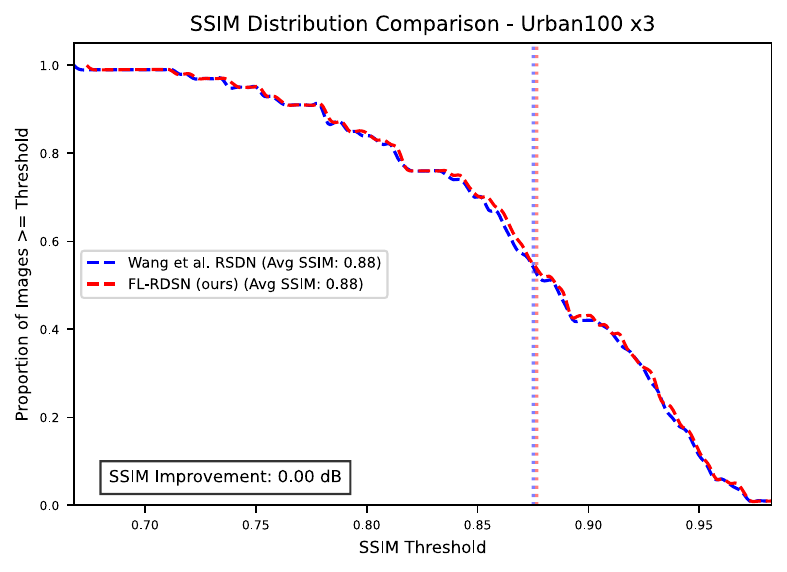}
\caption{Urban100, SSIM, \(\varsigma = 3\)}
\label{fig:urban100-3-ssim}
\end{subfigure}
\hfill
\begin{subfigure}{0.32\textwidth}
\includegraphics[width=\textwidth,height=5cm]{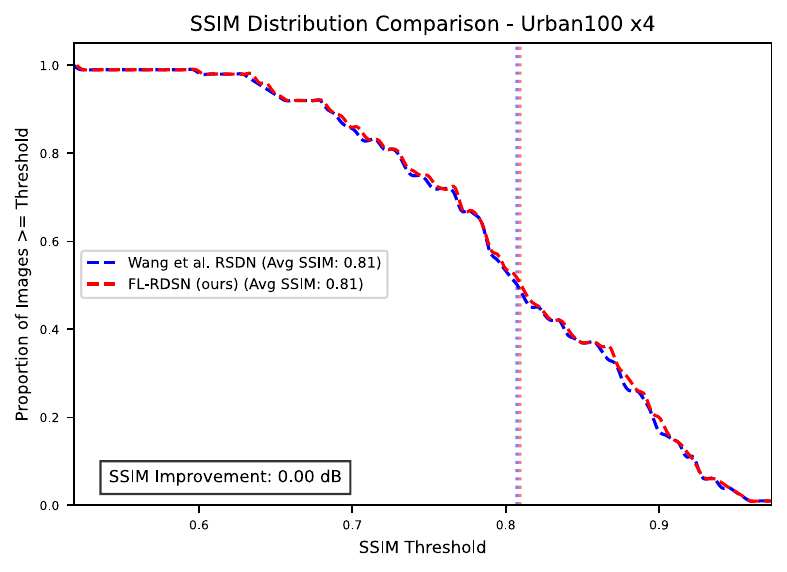}
\caption{Urban100, SSIM, \(\varsigma = 4\)}
\label{fig:urban100-4-ssim}
\end{subfigure}

\caption{Performance comparison for SSIM between State-of-the-Art and our FL-RDSN}
\label{fig:ssim-comparison}
\end{figure*}

% =================================

% ==================================
% =======================================
\subsection{Reconstructed Samples}
\label{samples}
In this section, we demonstrate the reconstructed samples. The figures are from DIV2K \cite{Agustsson_2017_CVPR_Workshops} testing sets. The scaling factor we choose is 2 because the reconstructed super-resolution images are less distinguishable from the low-resolution images compared to the scaling factor equal to 3 or 4. But if one can distinguish them, it means our model performs well.
% ============pics================
\begin{figure}[H]
    \centering
    \begin{subfigure}{0.49\columnwidth}
    \includegraphics[width=\textwidth, height=4cm]{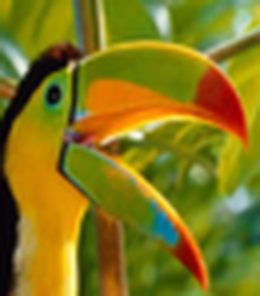}
    
    \caption{LR Img, \(\varsigma = 2\)}
    \label{fig:sub1}
    \end{subfigure}
    \hfill
    \begin{subfigure}{0.49\columnwidth}
    \includegraphics[width=\textwidth, height=4cm]{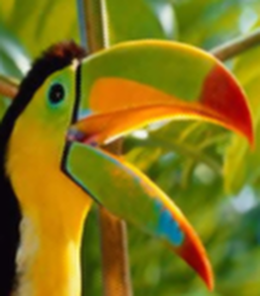}
    \caption{SR Img, \(\epsilon = 2.75\), \(\varsigma = 2\)}
    \label{fig:sub2}
    \end{subfigure}
    \begin{subfigure}{0.49\columnwidth}
    \includegraphics[width=\textwidth, height=4cm]{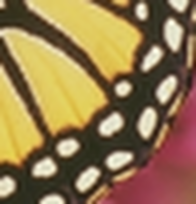}
    
    \caption{LR Img, \(\varsigma = 2\)}
    \label{fig:sub1}
    \end{subfigure}
    \hfill
    \begin{subfigure}{0.49\columnwidth}
    \includegraphics[width=\textwidth, height=4cm]{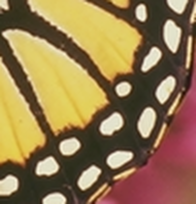}
    
    \caption{SR Img, \(\epsilon = 2.75\), \(\varsigma = 2\)}
    \label{fig:sub1}
    \end{subfigure}
    \hfill
    \begin{subfigure}{0.49\columnwidth}
    \includegraphics[width=\textwidth, height=4cm]{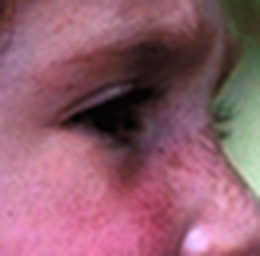}
    
    \caption{LR Img, \(\varsigma = 2\)}
    \label{fig:sub1}
    \end{subfigure}
    \hfill
    \begin{subfigure}{0.49\columnwidth}
    \includegraphics[width=\textwidth, height=4cm]{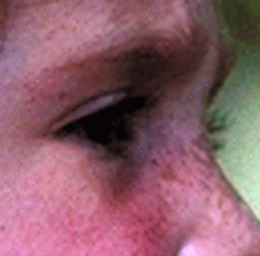}
    
    \caption{SR Img, \(\epsilon = 2.75\), \(\varsigma = 2\)}
    \label{fig:sub1}
    \end{subfigure}
    
    \hfill
    \caption{Reconstruction samples using PPFL-RDSN, scaling factor = 2}
    \label{example}
\end{figure}
\end{document}